\documentclass{article}

\PassOptionsToPackage{numbers}{natbib}
\usepackage[preprint]{neurips_2026}

\usepackage[utf8]{inputenc}
\usepackage[T1]{fontenc}

\usepackage{booktabs}
\usepackage{amsfonts}
\usepackage{nicefrac}
\usepackage{microtype}
\usepackage{xcolor}
\usepackage{colortbl}
\usepackage{graphicx}
\usepackage{amsmath}
\usepackage{amssymb}
\usepackage{enumitem}
\usepackage{algorithm}
\usepackage{algpseudocode}
\usepackage{tcolorbox}
\tcbuselibrary{skins, breakable}
\usepackage{tikz}
\usepackage{titletoc}
\usepackage{etoc}
\usepackage{alltt}

\usepackage{hyperref}

\newcommand{\runtimefield}[1]{\textcolor{salienceorange}{\texttt{#1}}}

\usetikzlibrary{arrows.meta, positioning, shapes.geometric, fit, backgrounds}

\definecolor{bestcolor}{RGB}{198, 224, 255}
\definecolor{hlcolor}{RGB}{255, 255, 180}
\definecolor{boxbg}{RGB}{245, 247, 250}
\definecolor{boxframe}{RGB}{180, 190, 205}
\definecolor{successgreen}{RGB}{30, 120, 50}
\definecolor{failred}{RGB}{180, 40, 40}
\definecolor{salienceorange}{RGB}{200, 130, 30}
\definecolor{successcolor}{RGB}{220, 245, 220}
\definecolor{failcolor}{RGB}{250, 225, 225}
\definecolor{saliencecolor}{RGB}{255, 243, 220}

\newcommand{\hl}[1]{\colorbox{hlcolor}{#1}}

\newtcolorbox{exbox_q}[1]{
  enhanced,
  colback=boxbg, colframe=boxframe,
  fonttitle=\bfseries\small, title={#1},
  boxrule=0.5pt, arc=2pt,
  left=5pt, right=5pt, top=1pt, bottom=1pt
}

\newtcolorbox{exbox}[1]{
  enhanced,
  breakable,
  colback=boxbg, colframe=boxframe,
  fonttitle=\bfseries\small, title={#1},
  boxrule=0.5pt, arc=2pt,
  left=5pt, right=5pt, top=3pt, bottom=3pt
}

\newtcolorbox{successbox}[1]{
  enhanced, colback=white, colframe=white,
  borderline west={3pt}{0pt}{green!60!black},
  fonttitle=\bfseries\small, title={#1},
  boxrule=0.3pt, arc=0pt,
  left=6pt, right=4pt, top=2pt, bottom=2pt
}

\newtcolorbox{failbox}[1]{
  enhanced, colback=white, colframe=white,
  borderline west={3pt}{0pt}{red!60!black},
  fonttitle=\bfseries\small, title={#1},
  boxrule=0.3pt, arc=0pt,
  left=6pt, right=4pt, top=2pt, bottom=2pt
}

\newtcolorbox{saliencebox}[1]{
  enhanced, colback=white, colframe=white,
  borderline west={3pt}{0pt}{orange!80!black},
  fonttitle=\bfseries\small, title={#1},
  boxrule=0.3pt, arc=0pt,
  left=6pt, right=4pt, top=2pt, bottom=2pt
}

\title{Reflective Prompted Policy Optimization: Trajectory-Grounded Revision and Salience Bias}
\author{%
\textbf{Rahaf Abu Hara},
\textbf{Vaibbhav Murarri},
\textbf{Claudio Zito} \\
Laboratory of AI and Robotics Research (LAIRR) \\
Heriot-Watt University \\
Dubai, United Arab Emirates \\
{\texttt{\{rha4001,vm81,C.Zito\}@hw.ac.uk}}%
}

\begin{document}

\maketitle

\begin{abstract}
Existing LLM-based policy optimizers see only scalar rewards: that a policy scored 0.45, but not whether the agent got stuck in a loop, fell into a hole on the third step, or performed well on 19 out of 20 rollouts and failed catastrophically on one. We propose Reflective Prompted Policy Optimization (R2PO), a two-stage LLM framework for policy search over compact policy classes that augments scalar reward feedback with trajectory-level behavioral evidence. A \emph{Search-LLM} acts as a global policy optimizer and proposes candidate policy parameters; the environment executes them; a \emph{Critic-LLM} then inspects the resulting rollouts and proposes targeted parameter revisions grounded in observed states, actions, and rewards. Across ten environments, ablations show R2PO's gains arise from a design that explicitly separates global search from behavior-grounded revision and uses selection to filter high-variance edits. We further identify a dominant failure mode, \emph{salience bias}. When presented with multiple rollouts, the Critic-LLM fixates on improving a single failure even when most trajectories succeed. In a three-trajectory variant, where the Critic-LLM is shown the best, worst, and median rollout from each evaluation, this behavior explains 76.6\% of regressions on CartPole.
R2PO mitigates this by reasoning over aggregate rollout statistics, median-trajectory selection, and a revision rule. Using a relatively small open-weight 20B-parameter model, R2PO achieves the highest mean best reward across all ten environments, while
reaching near-optimal performance substantially earlier in training (e.g., near-maximum CartPole reward within $\sim$500 episodes), and training far more stably than both deep RL and prior LLM-based methods. Together, these results show that treating trajectories as first-class in-context evidence, rather than background artifacts reduced to scalar returns, fundamentally changes how even comparatively small LLMs can search over policy spaces, enabling them to learn faster, diagnose more precisely, and reliably improve external controllers rather than tune them by trial and error.

\end{abstract}

\section{Introduction}
\label{sec:introduction}
 
Reinforcement learning has made substantial progress on policy optimization, but standard algorithms require careful hyperparameter tuning and millions of environment interactions to converge, with performance sensitive to implementation details. Recent work has shown that large language models can serve as black-box optimizers for policy search, consuming histories of parameter-reward pairs and proposing new candidate policies through in-context learning~\citep{yang2024opro,zhou2025prompted}. These methods treat the LLM as an optimizer over the policy parameter space, requiring no task-specific algorithmic design. 

 However, existing LLM-based policy optimizers operate on scalar reward feedback alone. The optimizer sees \emph{that} a policy scored 0.45, but not \emph{why}: whether the agent got stuck in a loop, fell into a hole on the third step, or performed well on 19 out of 20 rollouts but catastrophically failed on one. Scalar feedback provides no diagnostic signal about what went wrong, limiting the optimizer to trial-and-error tuning rather than targeted correction.

We introduce \textbf{Reflective Prompted Policy Optimization (R2PO)}, a two-stage LLM framework for policy search over compact policy classes (linear policies for continuous control and tabular policies for discrete tasks). A \emph{Search-LLM} acts as a global policy optimizer: given the history of evaluated policies and their scalar rewards, it proposes a candidate policy parameter vector. The candidate is executed in the environment, producing rollout trajectories that capture states, actions, and rewards. A \emph{Critic-LLM} then inspects this evidence and proposes a targeted parameter revision. Finally, a selection step compares the original and revised candidates and keeps whichever achieves the higher reward, limiting damage from over-corrections or LLM hallucinations.

Trajectory-grounded diagnosis is a powerful tool that enables better decisions than scalar-only feedback, but it can also cause sharp regressions when the trajectories used for diagnosis are unrepresentative. We identify a recurring failure pattern, \emph{salience bias}: when the Critic-LLM is shown multiple trajectories from the same policy, it treats the worst-case trajectory as representative, even when the policy is working well. We mitigate salience bias by feeding the Critic-LLM a more representative view of policy behavior: (i) aggregate rollout statistics that summarize overall performance, (ii) a median trajectory that reflects typical rather than worst-case behavior, and (iii) a revision rule that preserves policies that already perform well.

Our contributions are:
\begin{itemize}[leftmargin=*,itemsep=2pt,topsep=2pt]

\item \textbf{R2PO}, a two-stage LLM policy-optimization framework that combines scalar reward search with trajectory-grounded behavioral diagnosis, treating rollouts as first-class in-context evidence.

\item \textbf{A systematic ablation} across ten environments over search budget, prompt design, two-stage architecture, selection, and trajectory evidence, showing that R2PO's gains cannot be explained by more LLM calls or richer prompts alone, but require the combination of trajectory-grounded revision, two-stage design, and keep-best selection.

\item \textbf{A mechanistic analysis} identifying and quantifying a dominant failure mode of trajectory-grounded LLM revision, \emph{salience bias}, in which the Critic-LLM overweights vivid worst-case rollouts when multiple trajectories are available, accounting for 76.6\% of regressions in a three-trajectory setting on CartPole.

\item \textbf{Empirical validation} using a 20B open-weight LLM: R2PO achieves the highest mean reward on 9 of 10 environments and ties on the 10th, matches or exceeds all baselines on mean best reward, reaches near-optimal performance substantially earlier in training, and trains more stably than both deep RL and prior LLM-based baselines.\footnote{Code is available at: \url{https://github.com/HWUDLabAIRoboticsResearch/R2PO}}
\end{itemize}

\section{Related Work}
\label{sec:related_work}

Recent work integrates LLMs into RL pipelines in two main ways. One line uses LLMs to shape components of the RL loop, such as reward functions, action selection, or auxiliary policy guidance, while leaving optimization to conventional RL algorithms~\citep{kwon2023reward,ma2024dreureka,han2024flightcontrol,wu2024qtable,zhou2024policyteacher}. In these approaches, the LLM acts as an auxiliary module rather than as the optimizer itself.

A second line places the LLM at the center of optimization. OPRO~\citep{yang2024opro} uses LLMs to optimize arbitrary objective functions through iterative prompting; ProPS~\citep{zhou2025prompted} extends this to RL by consuming parameter-reward histories and proposing policy updates directly. Our work operates in this regime, but differs fundamentally: rather than optimizing from scalar feedback alone, we introduce a second LLM stage that reasons over the proposed policy's actual rollout behavior, enabling targeted diagnosis of failure modes that scalar rewards cannot reveal.

Self-improvement and reflection methods for LLMs are also related. Self-Refine~\citep{madaan2023selfrefine} performs iterative generate-critique-revise loops for single outputs; Reflexion~\citep{shinn2023reflexion} extends this to agents by converting task feedback into verbal reflections stored in episodic memory. In Reflexion, the LLM \emph{is} the policy and iteratively retries the same task using verbal reflections from prior attempts. 
%In R2PO, the LLM is an \emph{optimizer} of separate policy parameters: the Critic-LLM is stateless, diagnoses an external policy's behavior, and produces a concrete parameter vector that competes with the original through selection rather than free-form text.
In R2PO, the LLM is an \emph{optimizer} of separate policy parameters: the Critic-LLM diagnoses an external policy's rollout behavior and produces a concrete parameter vector evaluated against the original.

% A recent direction uses trajectories as supervision for LLM \emph{training}. TreeRL~\citep{hou2025treerl} adds on-policy tree search with process supervision; TSR~\citep{djuhera2026tsr} constructs higher-quality training trajectories through lightweight tree-style search before optimization. These methods optimize LLM weights via gradient updates. R2PO instead uses rollout trajectories as \emph{in-context evidence} to revise external policies, with no gradient computation or fine-tuning. The trajectories serve as diagnostic input to the LLM, not as a training signal for the LLM itself.

A recent direction uses trajectories as supervision for LLM \emph{training} via gradient updates: TreeRL~\citep{hou2025treerl} adds on-policy tree search with process supervision, and TSR~\citep{djuhera2026tsr} constructs higher-quality training trajectories through lightweight tree-style search. R2PO instead uses rollout trajectories as \emph{in-context evidence} to revise external policies, with no gradient computation or fine-tuning.

\section{Methodology}
\label{sec:methodology}

\subsection{Problem setting}
\label{sec:problem-setting}
 
We consider episodic policy search with parameterized policies $\pi_\theta$ and the objective of maximizing expected return $J(\theta) = \mathbb{E}[R(\tau)]$, where $\tau$ is a trajectory generated by $\pi_\theta$. We focus on compact policy classes whose parameter vectors fit within an LLM context window: linear policies $\pi_\theta(s) = \theta^\top \phi(s)$ for continuous-state tasks ($\arg\max$
over action logits for discrete actions) and tabular policies $\pi_\theta(s) = \theta_s$, with $\theta_s \in \mathcal{A}$, for discrete-state tasks. Extending R2PO to high-dimensional neural policies is an important direction for future work (Section~\ref{sec:discussion}).

Given a parameter vector $\theta$, we evaluate it by running $K$ independent environment rollouts and averaging their returns. We denote this evaluation
\begin{equation*}
\textsc{Eval}(\theta) \rightarrow (\bar{R}, \mathcal{T}),
\end{equation*}
where $\bar{R}$ is the mean return and $\mathcal{T} = \{\tau_1, \ldots, \tau_K\}$ is the set of recorded trajectories.

\subsection{The R2PO framework}
\label{sec:R2PO-framework}
 
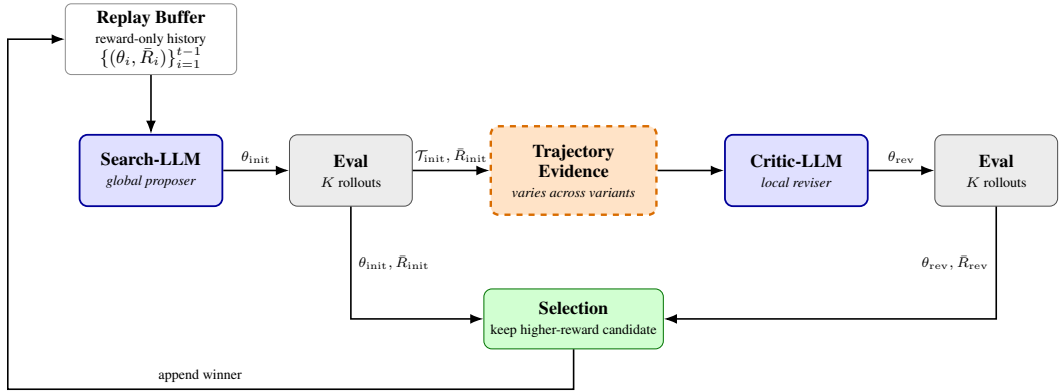
\begin{figure}[ht]
\centering
\resizebox{\textwidth}{!}{%
\begin{tikzpicture}[
  >=Latex,
  every node/.style={font=\small},
  llm/.style={rectangle, draw=blue!60!black, fill=blue!12, rounded corners=4pt,
              minimum width=2.5cm, minimum height=1.25cm, align=center, line width=0.9pt},
  env/.style={rectangle, draw=gray!60!black, fill=gray!15, rounded corners=4pt,
              minimum width=2.15cm, minimum height=1.25cm, align=center},
  evi/.style={rectangle, draw=orange!85!black, fill=orange!22, rounded corners=4pt,
              minimum width=2.9cm, minimum height=1.55cm, align=center, line width=1.2pt, dashed},
  sel/.style={rectangle, draw=green!45!black, fill=green!18, rounded corners=4pt,
              minimum width=2.8cm, minimum height=1.05cm, align=center},
  hist/.style={rectangle, draw=black!45, fill=white, rounded corners=3pt,
               minimum width=3cm, minimum height=0.95cm, align=center}
]

% Top row
\node[llm] (search) at (0,0) {\textbf{Search-LLM}\\\scriptsize\emph{global proposer}};
\node[env] (evala) at (3.5,0) {\textbf{Eval}\\\scriptsize $K$ rollouts};
\node[evi] (evi) at (7.4,0) {\textbf{Trajectory}\\\textbf{Evidence}\\\scriptsize\emph{varies across variants}};
\node[llm] (critic) at (11.3,0) {\textbf{Critic-LLM}\\\scriptsize\emph{local reviser}};
\node[env] (evalb) at (14.8,0) {\textbf{Eval}\\\scriptsize $K$ rollouts};

% Replay buffer
\node[hist] (hist) at (0,2.3) {\textbf{Replay Buffer}\\\scriptsize reward-only history \\$\{(\theta_i,\bar{R}_i)\}_{i=1}^{t-1}$};

% Selection
\node[sel] (sel) at (7.4,-2.6) {\textbf{Selection}\\\scriptsize keep higher-reward candidate};

% Main horizontal arrows
\draw[->, thick] (hist.south) -- (search.north);
\draw[->, thick] (search) -- node[above, font=\scriptsize] {$\theta_{\mathrm{init}}$} (evala);
\draw[->, thick] (evala) -- node[above, font=\scriptsize] {$\mathcal{T}_{\mathrm{init}},\bar{R}_{\mathrm{init}}$} (evi);
\draw[->, thick] (evi) -- (critic);
\draw[->, thick] (critic) -- node[above, font=\scriptsize] {$\theta_{\mathrm{rev}}$} (evalb);

% Eval to selection arrows: straight routed, no curves
\draw[->, thick]
  (evala.south) |- node[pos=0.25, right, font=\scriptsize] {$\theta_{\mathrm{init}},\bar{R}_{\mathrm{init}}$} (sel.west);

\draw[->, thick]
  (evalb.south) |- node[pos=0.25, left, font=\scriptsize] {$\theta_{\mathrm{rev}},\bar{R}_{\mathrm{rev}}$} (sel.east);

% Selection back to replay buffer:
\coordinate (hist_in) at ([xshift=-1.0cm]hist.west);

\draw[->, thick]
  (sel.south) -- ++(0,-0.7) -| node[pos=0.33, above, font=\scriptsize, align=center] {append winner} (hist_in) -- (hist.west);

\end{tikzpicture}
}
\caption{\textbf{The two-stage R2PO framework.} The Search-LLM proposes candidate parameters $\theta_{\mathrm{init}}$ conditioned on a reward-only replay history $\{(\theta_i,\bar{R}_i)\}_{i=1}^{t-1}$ of previously evaluated parameters and their mean rewards. The environment evaluates $\theta_{\mathrm{init}}$ over $K$ rollouts, returning the mean reward and rollout trajectories from which \emph{Trajectory Evidence} (orange, dashed) is constructed. The Critic-LLM uses this evidence to propose a revised vector $\theta_{\mathrm{rev}}$, which is re-evaluated; a selection step retains whichever candidate achieves the higher reward before appending it to the replay buffer. Trajectory Evidence is the central design choice we study: in R2PO, it consists of the median rollout, aggregate rollout statistics, and a revision rule applied when the policy is already strong.}
\label{fig:R2PO-framework}
\end{figure}

R2PO uses two LLMs in distinct roles (Figure~\ref{fig:R2PO-framework}). The Search-LLM is the \emph{global policy optimizer}: given the history of previously evaluated policies and their scalar mean rewards, it proposes where to move next in the parameter space. Scalar history suffices at this global level, where candidates are compared on a uniform ranking signal; trajectory-grounded reasoning is reserved for the local revision step. The Critic-LLM is the \emph{trajectory-based reviser}: given a calibrated summary of the behavior observed when the Search-LLM's proposal was executed, it inspects what went wrong and proposes a targeted revision. In practice, both roles use the \emph{same} underlying 20B model, but via two independent calls with distinct prompts tailored to their respective roles.

\paragraph{Stage 1: Search-LLM proposal.}
At iteration $t$, the Search-LLM is prompted with the optimization task and a reward-only history $H_t = \{(\theta_i, \bar{R}_i)\}_{i=1}^{t-1}$, and proposes an initial parameter vector $\theta_t^{(0)}$, which the environment evaluates as $(\bar{R}_t^{(0)}, \mathcal{T}_t^{(0)})$.

\paragraph{Stage 2: Critic-LLM reflection.}
The Critic-LLM is prompted with a natural-language environment description, the proposed parameters $\theta_t^{(0)}$, the same reward-only history $H_t$, and a calibrated trajectory evidence package built from the $K$ rollouts in $\mathcal{T}_t^{(0)}$. From this evidence, it proposes a revised parameter vector $\theta_t^{(1)}$. The evidence package has three components.

\emph{(i) Median trajectory.} From the $K$ rollouts, we select the rollout whose return is closest to the median of the $K$ returns (interpolated when $K$ is even). This provides a concrete example of typical policy behavior, avoiding both best-case and worst-case outliers.
 
\emph{(ii) Aggregate rollout statistics.} A structured statistical report computed over all $K$ rollouts: reward mean, minimum, and maximum; episode length mean, minimum, and maximum; and success and failure rates. These aggregates convey how often the policy fails and by how much without placing a specific failed execution trace in front of the Critic-LLM. 
% The report requires no domain-specific feature engineering: returns and episode lengths are universal across environments.

\emph{(iii) Revision rule.} We add an explicit abstention instruction to the Critic-LLM's prompt, controlled by a threshold $\tau_c$. If the initial mean reward $\bar{R}_t^{(0)}$ exceeds $\tau_c$ and the rollout statistics do not indicate a consistent failure pattern, the Critic-LLM is instructed to keep the parameters unchanged or make only a very small, strongly justified edit. The threshold $\tau_c$ is set per environment based on the observable reward scale (see Appendix~\ref{app:thresholds}).

% The two stages play complementary roles: the Search-LLM performs a global numerical search over the parameter space without any knowledge of what policies actually do, while the Critic-LLM performs a local diagnosis grounded in a specific execution evidence package that does not exist when the Search-LLM makes its proposal. The environment mediates between them.

\paragraph{Selection.}
We evaluate the revised parameters and keep whichever candidate performs better:
\begin{equation}
\theta_t =
\begin{cases}
\theta_t^{(1)} & \text{if } \bar{R}_t^{(1)} \geq \bar{R}_t^{(0)},\\
\theta_t^{(0)} & \text{otherwise.}
\end{cases}
\end{equation}

The selected parameters and their mean reward are appended to the history for subsequent iterations. Algorithm~\ref{alg:R2PO} in Appendix~\ref{app:algorithm} summarizes the full R2PO loop.

\section{Experiments}
\subsection{Experimental setup}
\label{sec:experimental-setup}

We evaluate on ten environments spanning discrete and continuous action spaces, as well as stochastic and deterministic dynamics: CartPole-v1, FrozenLake-v1, MountainCar-v0, MountainCarContinuous-v0, InvertedPendulum-v5, InvertedDoublePendulum-v5, and Swimmer-v5 \citep{towers2024gymnasium,todorov2012mujoco}, and three custom environments (Maze, Nim, and Pong). All LLM calls use \texttt{gpt-oss:20b-cloud}, an Ollama deployment of the OpenAI gpt-oss model family, with cloud-hosted inference \citep{openai2025gptoss,ollama_gptoss20b_cloud}. SB3 baseline hyperparameters are taken from the official Stable-Baselines3 library~\citep{raffin2021stable}, using RL Baselines3 Zoo~\citep{rl-zoo3} presets where available and library defaults otherwise. The revision threshold $\tau_c$ is set near each environment's strongest empirically achievable reward (Appendix~\ref{app:thresholds}). Appendix~\ref{app:tau-sensitivity} shows R2PO is robust to $\pm 50\%$ threshold variation and remains competitive without it.

We compare three method tiers. \emph{Baselines}: ProPS, ProPS+ \citep{zhou2025prompted}, and the best-performing Stable-Baselines3 (SB3)~\citep{raffin2021stable} algorithm per environment selected from  PPO \citep{schulman2017ppo}, TRPO \citep{schulman2015trpo}, SAC \citep{haarnoja2018soft}, DQN \citep{mnih2015human}, A2C \citep{mnih2016asynchronous}, DDPG \citep{lillicrap2016continuous}, and TD3 \citep{fujimoto2018addressing}. \emph{Decomposition ablations}: PureSearch, ActorSecondPass, CriticOnly, and AlwaysCritic, each isolating one factor that could explain R2PO's gains (Section~\ref{sec:ablation}). \emph{Trajectory-evidence variants}: R2PO, RepTraj , and ThreeTraj, each instantiating a different construction of trajectory evidence for the Critic-LLM (Section~\ref{sec:ablation}).

All methods are evaluated over 10 independent runs with $K{=}20$ rollouts per evaluation. Each method receives a matched budget of 200 LLM calls and 4{,}000 episodes per run: two-call methods (R2PO, RepTraj, ThreeTraj, AlwaysCritic, ActorSecondPass, PureSearch) run for $T{=}100$ iterations with 2 LLM calls and $2K{=}40$ rollouts per iteration; single-call methods (ProPS, ProPS+, CriticOnly) run for $2T{=}200$ iterations with 1 LLM call and $K$ rollouts per iteration. SB3 baselines train on 4{,}000 episodes per run. We report two metrics, both averaged over 10 runs with $\pm$ standard deviation: \emph{mean reward} (the average reward across all iterations) and \emph{mean best reward} (mean of the best reward achieved during training).

\subsection{Main results}
\label{sec:main-results}

% Table~\ref{tab:main-mean} reports the mean reward across 10 seeds for R2PO, ProPS, ProPS+, and the best SB3 baseline per environment. Table~\ref{tab:main-best} reports the mean best reward. R2PO achieves the highest mean reward on 9 out of 10 environments. The exception is MountainCarContinuous, where SAC achieves $82.33 \pm 2.44$ versus R2PO's $81.61 \pm 4.91$ (essentially tied). R2PO outperforms ProPS+ on all 10 environments, often by large margins: on InvertedPendulum, $756$ vs.\ $310$; on CartPole, $475$ vs.\ $253$; on Swimmer, $260$ vs.\ $162$.

% On mean best reward, R2PO achieves the maximum possible return on CartPole ($500$), InvertedPendulum ($1000$), Pong ($3.0$), and Nim ($1.0$), matching or exceeding the best SB3 baseline on all 10 environments.

% paragraph 1 + 2:
Table~\ref{tab:main-best} shows that the best policies found by R2PO match or exceed all baselines on all 10 environments, reaching the maximum possible return on CartPole ($500$), InvertedPendulum ($1000$), Pong ($3.0$), and Nim ($1.0$). 
R2PO further achieves the highest mean reward across 9 of 10 environments (Table~\ref{tab:main-mean}); the only exception is MountainCarContinuous, where SAC slightly leads ($82.33 \pm 2.57$ vs our $81.61 \pm 5.18$). Compared to ProPS+, R2PO performs better on all 10 environments, often by large margins, and Appendix~\ref{sec:significance-tests} shows that these gains are statistically significant.

\begin{figure*}[t]
\centering
\includegraphics[width=\textwidth]{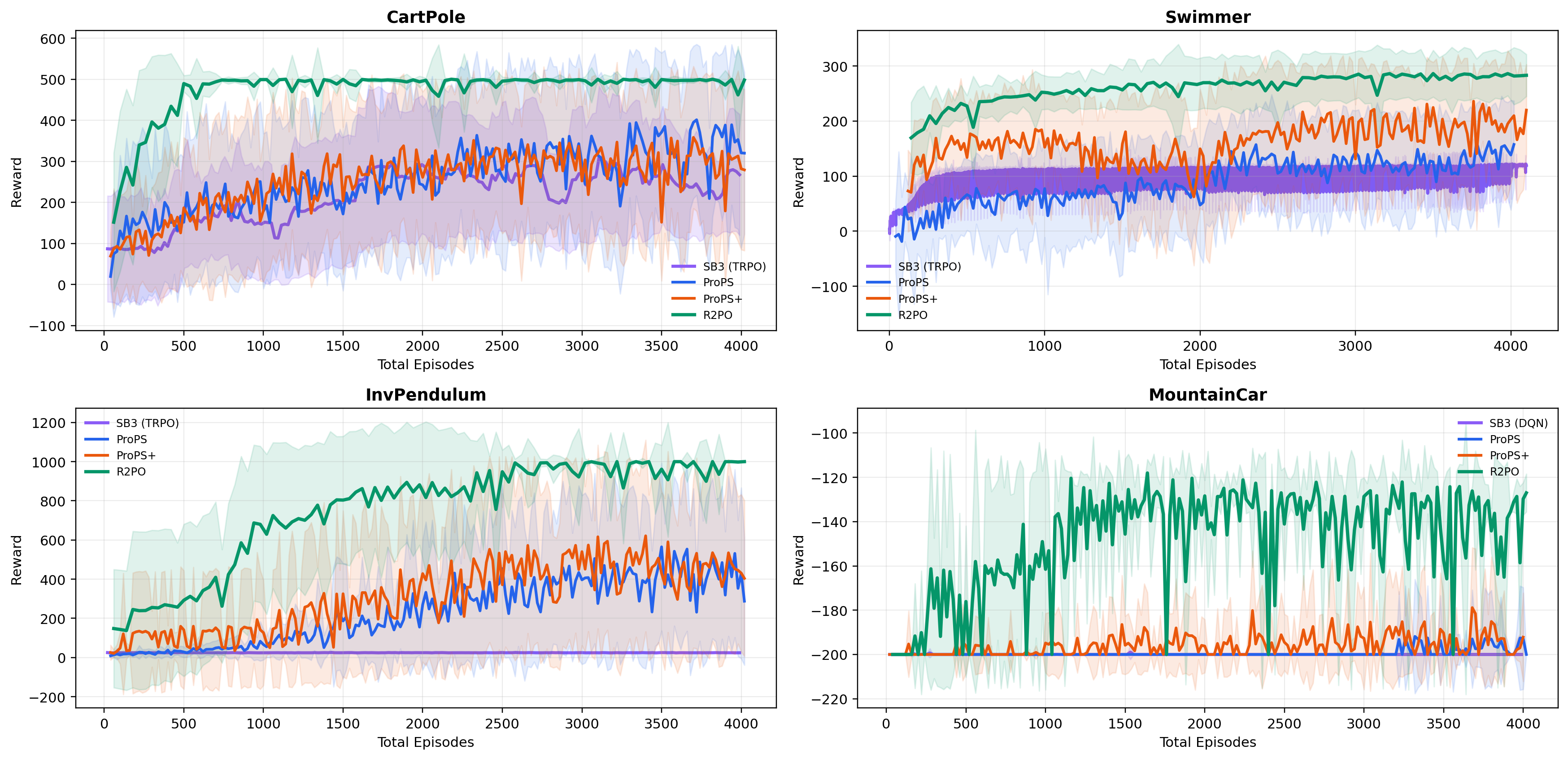}
\caption{Learning curves (mean $\pm$ standard deviation over 10 independent runs) for R2PO, ProPS, ProPS+, and the best SB3 baseline on four representative environments. R2PO reaches strong performance earlier and maintains it more consistently than other baselines. Curves for the remaining six environments are in Appendix~\ref{app:learning-curves} and show the same overall pattern.}
\label{fig:learning-curves}
\end{figure*}

Beyond peak performance, R2PO maintains strong policies once they are found: its mean rewards remain close to its best rewards (e.g., CartPole $\sim$475/500; InvertedPendulum $\sim$756/1000), indicating stable training rather than isolated spikes (Appendix~\ref{app:stability}). Prior LLM-based methods are less stable (e.g., ProPS+ on InvertedPendulum reaches a best of $\sim$658 but averages only $\sim$310, and on MountainCarContinuous reaches $\sim$99 best vs.\ $\sim$18 mean). R2PO also learns faster, achieving near-optimal CartPole performance within $\sim$500 episodes (vs.\ $\sim$4{,}000 for ProPS+ and SB3) and surpassing all baselines on Swimmer within $\sim$1{,}000 episodes. Figure~\ref{fig:learning-curves} shows learning curves on four representative environments; full curves for all ten are in Appendix~\ref{app:learning-curves}.

Finally, Appendix~\ref{app:props-comparison} reports an ablation-style comparison against the published ProPS results~\citep{zhou2025prompted}, which were obtained with \emph{double} the interaction budget (8{,}000 episodes) and a larger proprietary model (GPT-4o). Even under these conditions, R2PO---trained with 4{,}000 episodes and an open-weight GPT-oss:20B model---still matches or exceeds ProPS+ and the best SB3 baseline on 9 of 10 environments.

% ── Main Results: Mean Reward ──
\begin{table}[t]
\centering
\caption{Mean reward ($\pm$ standard deviation) across 10 independent runs, averaged across all training iterations. \colorbox{bestcolor}{\textbf{Best}} per environment is highlighted. R2PO achieves the highest mean reward on 9 of 10 environments and is statistically tied with SAC on MountainCarContinuous ($81.61 \pm 5.18$ vs.\ $82.33 \pm 2.57$).}
\label{tab:main-mean}
\small
\setlength{\tabcolsep}{4pt}

\begin{tabular}{@{}lcccc@{}}
\toprule
Environment & ProPS & ProPS+ & Best SB3 & R2PO \\
\midrule
Nim & $-0.59 \pm 0.21$ & $-0.25 \pm 0.07$ & $0.01 \pm 0.32$ {\scriptsize(A2C)} & \cellcolor{bestcolor}$\mathbf{0.61 \pm 0.03}$ \\
Pong & $0.74 \pm 0.63$ & $1.22 \pm 0.58$ & $1.02 \pm 0.74$ {\scriptsize(PPO)} & \cellcolor{bestcolor}$\mathbf{2.51 \pm 0.21}$ \\
Swimmer & $89.22 \pm 48.68$ & $162.05 \pm 66.07$ & $44.60 \pm 7.30$ {\scriptsize(TRPO)} & \cellcolor{bestcolor}$\mathbf{260.35 \pm 36.05}$ \\
MountainCarCon & $-23.45 \pm 41.65$ & $17.90 \pm 37.48$ & \cellcolor{bestcolor}$\mathbf{82.33 \pm 2.57}$ {\scriptsize(SAC)} & $81.61 \pm 5.18$ \\
MountainCar & $-199.31 \pm 2.18$ & $-195.81 \pm 3.73$ & $-199.99 \pm 0.02$ {\scriptsize(DQN)} & \cellcolor{bestcolor}$\mathbf{-147.84 \pm 7.11}$ \\
InvDblPend & $79.44 \pm 16.50$ & $86.71 \pm 17.70$ & $86.04 \pm 0.33$ {\scriptsize(TRPO)} & \cellcolor{bestcolor}$\mathbf{158.51 \pm 68.41}$ \\
InvPendulum & $234.14 \pm 169.76$ & $309.52 \pm 247.92$ & $24.35 \pm 0.13$ {\scriptsize(TRPO)} & \cellcolor{bestcolor}$\mathbf{756.08 \pm 154.50}$ \\
FrozenLake & $0.05 \pm 0.05$ & $0.37 \pm 0.06$ & $0.02 \pm 0.02$ {\scriptsize(TRPO)} & \cellcolor{bestcolor}$\mathbf{0.62 \pm 0.07}$ \\
CartPole & $258.09 \pm 138.58$ & $253.06 \pm 133.35$ & $216.92 \pm 63.34$ {\scriptsize(TRPO)} & \cellcolor{bestcolor}$\mathbf{474.67 \pm 16.90}$ \\
Maze & $-1.03 \pm 0.18$ & $0.76 \pm 0.05$ & $\mathbf{0.83 \pm 0.13}$ {\scriptsize(A2C)} & \cellcolor{bestcolor}$\mathbf{0.83 \pm 0.06}$ \\
\bottomrule
\end{tabular}
\end{table}

% \vspace{-10}
 
% ── Main Results: Mean Best Reward ──
\begin{table}[t]
\centering
\caption{Mean best reward ($\pm$ standard deviation) across 10 independent runs. \colorbox{bestcolor}{\textbf{Best}} per environment is highlighted. R2PO achieves the strongest peak performance on all environments.}
\label{tab:main-best}
\small
\setlength{\tabcolsep}{4pt}
\begin{tabular}{@{}lcccc@{}}
\toprule
Environment & ProPS & ProPS+ & Best SB3 & R2PO \\
\midrule
Nim & $0.75 \pm 0.30$ & \cellcolor{bestcolor}$\mathbf{1.00 \pm 0.00}$ & $0.88 \pm 0.27$ {\scriptsize(A2C)} & \cellcolor{bestcolor}$\mathbf{1.00 \pm 0.00}$ \\
Pong & $2.15 \pm 0.95$ & $2.78 \pm 0.41$ & $2.33 \pm 0.78$ {\scriptsize(PPO)} & \cellcolor{bestcolor}$\mathbf{3.00 \pm 0.00}$ \\
Swimmer & $208.52 \pm 68.38$ & $274.86 \pm 38.81$ & $67.89 \pm 20.92$ {\scriptsize(TRPO)} & \cellcolor{bestcolor}$\mathbf{294.57 \pm 43.79}$ \\
MountainCarCon & $75.37 \pm 30.57$ & $98.70 \pm 0.50$ & $94.81 \pm 0.42$ {\scriptsize(SAC)} & \cellcolor{bestcolor}$\mathbf{98.75 \pm 0.44}$ \\
MountainCar & $-191.84 \pm 25.80$ & $-150.21 \pm 26.62$ & $-197.47 \pm 3.66$ {\scriptsize(DQN)} & \cellcolor{bestcolor}$\mathbf{-111.04 \pm 3.92}$ \\
InvDblPend & $112.18 \pm 20.94$ & $128.81 \pm 54.51$ & $98.25 \pm 2.56$ {\scriptsize(TRPO)} & \cellcolor{bestcolor}$\mathbf{254.04 \pm 232.39}$ \\
InvPendulum & $649.91 \pm 432.17$ & $657.88 \pm 444.42$ & $28.49 \pm 0.71$ {\scriptsize(TRPO)} & \cellcolor{bestcolor}$\mathbf{1000.00 \pm 0.00}$ \\
FrozenLake & $0.24 \pm 0.14$ & $0.90 \pm 0.04$ & $0.12 \pm 0.12$ {\scriptsize(TRPO)} & \cellcolor{bestcolor}$\mathbf{0.93 \pm 0.05}$ \\
CartPole & $427.74 \pm 143.04$ & $396.21 \pm 155.21$ & $490.76 \pm 29.22$ {\scriptsize(TRPO)} & \cellcolor{bestcolor}$\mathbf{500.00 \pm 0.00}$ \\
Maze & $-0.70 \pm 0.88$ & \cellcolor{bestcolor}$\mathbf{0.97 \pm 0.00}$ & \cellcolor{bestcolor}$\mathbf{0.97 \pm 0.00}$ {\scriptsize(A2C)} & \cellcolor{bestcolor}$\mathbf{0.97 \pm 0.00}$ \\
\bottomrule
\end{tabular}

\end{table}

\subsection{Ablation study and decomposition}
\label{sec:ablation}
R2PO combines (i) a two-stage LLM pipeline, (ii) trajectory-grounded behavioral evidence, (iii) a reward-only replay history, (iv) a revision rule, and (v) a keep-best selection mechanism. We ask two questions: \emph{(1) which components are actually necessary for the observed gains?} and \emph{(2) given a two-stage architecture, how should rollout evidence be summarized for the Critic-LLM?} We address (1) using architectural ablations (Section~\ref{sec:decomposition}) and (2) using trajectory-evidence variants that differ only in what the Critic-LLM sees (Section~\ref{sec:traj-representation}).
Table~\ref{tab:variants} reports mean reward across all ten environments; mean best reward (Appendix~\ref{app:variants-best}) shows the same overall pattern.

\subsubsection{Decomposition: are all components of R2PO necessary?}
\label{sec:decomposition}

We design four ablation variants, each removing or altering a single aspect of R2PO.

% \paragraph{Extra budget vs.\ architecture.}
\textbf{Is it just extra budget?}
\textbf{PureSearch} draws two proposals from the same Search-LLM context and keeps the better one, doubling the search budget without structural change. \textbf{ActorSecondPass} runs the scalar-only optimizer twice in sequence: the second call sees the first call's result and reward but no trajectory information. Both variants test whether R2PO's gains reduce to additional LLM calls or a generic sequential structure. Both are consistently among the weakest methods, failing systematically on Maze and Nim. Extra budget, whether parallel or sequential, does not explain R2PO's gains.

% \paragraph{Prompt richness vs.\ two-stage design.}
\textbf{Is it just the richer prompt?}
\textbf{CriticOnly} replaces the two-stage loop with a single rich Critic-style prompt that internally performs draft--critique--revise within one LLM call, receiving scalar history and environment description but never actual rollouts. This isolates prompt design from the effect of having a separate reviser that reasons over realized behavior. CriticOnly is the strongest single-factor variant, yet R2PO outperforms it on all 10 environments, with large gaps on Swimmer ($\sim$260 vs.\ $\sim$55) and InvertedPendulum ($\sim$756 vs.\ $\sim$127). A richer prompt helps but cannot substitute for a two-stage design in which the reviser accesses execution behavior.

% \paragraph{Role of keep-best selection.}
\textbf{Is selection necessary?}
\textbf{AlwaysCritic} is identical to the two-stage variant RepTraj (Section~\ref{sec:traj-representation}, which uses a single representative rollout as trajectory evidence), except that it always commits the Critic-LLM's revision without selection. RepTraj outperforms AlwaysCritic on all 10 environments, often by large margins (InvertedPendulum: $\sim$380 vs.\ $\sim$141; MountainCarContinuous: $\sim$$-94$ vs.\ $\sim$$-131$). With architecture and evidence held constant, committing all revisions unconditionally introduces large variance and frequent regressions. We investigate why trajectory evidence fails to help when revisions are unconditional in Section~\ref{sec:mechanistic}, showing that it reflects high-variance, trajectory-conditioned edits often based on unrepresentative rollouts.

\textbf{Summary.}
No single ablation recovers R2PO's performance. Extra LLM calls (PureSearch, ActorSecondPass), richer prompts alone (CriticOnly), and two-stage reasoning without selection (AlwaysCritic) are each insufficient. R2PO's gains arise from the \emph{combination} of (i) a two-stage architecture separating global proposal from local revision, (ii) trajectory-grounded reasoning in the reviser, and (iii) a keep-best selection rule that filters destructive edits.

\subsubsection{Trajectory representation: how should the Critic-LLM see rollouts?}
\label{sec:traj-representation}

With the two-stage architecture and selection rule fixed, the remaining design question is how to summarize $K$ rollouts into a compact representation that fits in context and remains interpretable to the Critic-LLM. Feeding all $K$ trajectories as raw traces would exceed typical context limits and, in our pilots, led to incoherent edits; we therefore focus on compact summaries that preserve the key behavioral patterns. Providing only the scalar mean discards the behavioral information that motivates R2PO. We compare three structured representations, which differ only in how they summarize $\mathcal{T}_t^{(0)}$.

\textbf{RepTraj: single representative trajectory.}
\textbf{RepTraj} selects the rollout whose return is closest to the mean across the $K$ rollouts and passes it as a raw trajectory trace, together with the full reward and trajectory history. The Critic-LLM is always asked to revise, with no revision rule. This tests the simplest use of trajectory evidence: one approximately typical example.

\textbf{ThreeTraj: best, worst, and median trajectories.}
\textbf{ThreeTraj} aims to improve representativeness by exposing the Critic-LLM to the range of behaviors observed: it passes the highest-, lowest-, and median-reward trajectories from the $K$ rollouts, with reward-only history $H_t$. The Critic-LLM is again always asked to revise. This tests whether richer within-policy evidence yields better corrections.

\textbf{R2PO: median trajectory + aggregate statistics + Revision rule.}
R2PO shows the Critic-LLM (i) the median-reward trajectory, (ii) aggregate statistics over $K$ rollouts (reward mean/min/max, length mean/min/max, success/failure counts), and (iii) a revision  instruction (threshold $\tau_c$) that preserves the policy when performance is already strong. Unlike RepTraj, R2PO does not show extreme rollouts; unlike ThreeTraj, it communicates failure frequency numerically rather than by foregrounding a worst-case trace.

\textbf{Empirical comparison.}
As Tables~\ref{tab:variants} and \ref{tab:variants-best} show, R2PO outperforms RepTraj on all 10 environments in mean reward. ThreeTraj achieves slightly higher mean reward than R2PO on 2 discrete stochastic environments (Nim and FrozenLake) but is clearly worse elsewhere, especially on continuous-control tasks (Swimmer: $\sim$$110$ vs.\ $\sim$$260$; InvertedPendulum: $\sim$$194$ vs.\ $\sim$$756$). Section~\ref{sec:mechanistic} shows why naive multi-trajectory conditioning can hurt: when rollouts disagree, the Critic-LLM over-focuses on vivid worst-case traces, a phenomenon we call \emph{salience bias}. R2PO's evidence design preserves the benefits of trajectory-grounded diagnosis while reducing this vulnerability.

% ── Variants Table: Mean Reward ──
\begin{table}[t]
\centering
\caption{Mean reward ($\pm$ standard deviation) across 10 independent runs for R2PO and the ablation variants, averaged across all training iterations. \colorbox{bestcolor}{\textbf{Best}} per environment is highlighted. BestBaseline denotes the best among ProPS, ProPS+, and all SB3 algorithms. R2PO achieves the strongest average performance on most environments, while ThreeTraj wins by a small margin on 2 environments (Nim and FrozenLake, both discrete stochastic).}
\label{tab:variants}
\small
\setlength{\tabcolsep}{2pt}
\resizebox{\textwidth}{!}{%
\begin{tabular}{@{}lcccccccc@{}}
\toprule
Environment & BestBaseline & PureSearch & ActorSecondPass & CriticOnly & AlwaysCritic & RepTraj & ThreeTraj & R2PO \\
\midrule
Nim & $0.01 \pm 0.32$ {\tiny(A2C)} & $-0.36 \pm 0.22$ & $-0.40 \pm 0.33$ & $0.16 \pm 0.05$ & $0.07 \pm 0.20$ & $0.52 \pm 0.13$ & \cellcolor{bestcolor}$\mathbf{0.63 \pm 0.10}$ & $0.61 \pm 0.03$ \\
Pong & $1.22 \pm 0.58$ {\tiny(ProPS+)} & $0.86 \pm 0.49$ & $1.03 \pm 0.82$ & $2.06 \pm 0.58$ & $1.86 \pm 0.55$ & $2.20 \pm 0.41$ & $1.83 \pm 0.79$ & \cellcolor{bestcolor}$\mathbf{2.51 \pm 0.21}$ \\
Swimmer & $162.05 \pm 66.07$ {\tiny(ProPS+)} & $131.48 \pm 78.68$ & $111.89 \pm 96.37$ & $55.19 \pm 58.55$ & $111.45 \pm 54.50$ & $170.04 \pm 68.11$ & $109.56 \pm 75.57$ & \cellcolor{bestcolor}$\mathbf{260.35 \pm 36.05}$ \\
MountainCarCon & \cellcolor{bestcolor}$\mathbf{82.33 \pm 2.57}$ {\tiny(SAC)} & $-21.59 \pm 73.21$ & $-124.66 \pm 218.47$ & $-85.08 \pm 150.64$ & $-130.71 \pm 179.28$ & $-93.85 \pm 138.50$ & $-29.88 \pm 54.38$ & $81.61 \pm 5.18$ \\
MountainCar & $-195.81 \pm 3.73$ {\tiny(ProPS+)} & $-186.38 \pm 20.21$ & $-195.73 \pm 13.50$ & $-187.29 \pm 7.87$ & $-188.07 \pm 9.73$ & $-169.98 \pm 20.36$ & $-162.48 \pm 19.64$ & \cellcolor{bestcolor}$\mathbf{-147.84 \pm 7.11}$ \\
InvDblPend & $86.71 \pm 17.70$ {\tiny(ProPS+)} & $86.90 \pm 25.54$ & $88.17 \pm 8.98$ & $128.87 \pm 32.72$ & $99.11 \pm 21.14$ & $132.96 \pm 47.18$ & $100.07 \pm 24.90$ & \cellcolor{bestcolor}$\mathbf{158.51 \pm 68.41}$ \\
InvPendulum & $309.52 \pm 247.92$ {\tiny(ProPS+)} & $466.08 \pm 357.96$ & $480.08 \pm 324.97$ & $126.63 \pm 123.89$ & $141.41 \pm 228.65$ & $380.19 \pm 213.27$ & $194.25 \pm 235.22$ & \cellcolor{bestcolor}$\mathbf{756.08 \pm 154.50}$ \\
FrozenLake & $0.37 \pm 0.06$ {\tiny(ProPS+)} & $0.08 \pm 0.06$ & $0.07 \pm 0.04$ & $0.31 \pm 0.10$ & $0.31 \pm 0.08$ & $0.47 \pm 0.20$ & \cellcolor{bestcolor}$\mathbf{0.63 \pm 0.06}$ & $0.62 \pm 0.07$ \\
CartPole & $258.09 \pm 138.58$ {\tiny(ProPS)} & $337.72 \pm 156.62$ & $305.04 \pm 123.88$ & $418.23 \pm 45.36$ & $377.65 \pm 85.88$ & $416.15 \pm 88.71$ & $432.08 \pm 59.56$ & \cellcolor{bestcolor}$\mathbf{474.67 \pm 16.90}$ \\
Maze & $0.83 \pm 0.13$ {\tiny(A2C)} & $-1.10 \pm 0.03$ & $-1.01 \pm 0.33$ & $0.68 \pm 0.10$ & $0.74 \pm 0.07$ & $0.77 \pm 0.06$ & $0.82 \pm 0.12$ & \cellcolor{bestcolor}$\mathbf{0.83 \pm 0.06}$ \\
\bottomrule
\end{tabular}

}
\end{table}
 
\subsection{Mechanistic analysis}
\label{sec:mechanistic}

The ablations raise a key question: \emph{why} is trajectory-grounded revision so high-variance that it needs both selection and calibration? AlwaysCritic and CriticOnly perform similarly even though AlwaysCritic sees more information, and ThreeTraj, which sees \emph{three} rollouts, does worse. To investigate, we focus on CartPole and Pong—where CriticOnly outperforms AlwaysCritic—and use AlwaysCritic as our main diagnostic, since it isolates trajectory-based revision without selection or calibration. For each environment, we analyze 1{,}000 revision decisions (10 runs, 100 iterations).

\textbf{Trajectory-grounded diagnosis is high-variance.}
On Pong, 31.6\% of AlwaysCritic iterations are regressions; on CartPole, 34\%. The same diagnostic mechanism, the Critic-LLM reading a trajectory, identifying a failure point, and making a fix, produces both the largest improvements and regressions.

\textbf{The root cause: unrepresentative trajectories.}
Under AlwaysCritic, the Critic-LLM sees only one rollout out of $K{=}20$. In stochastic environments, a single rollout is a \emph{sample} of policy behavior, not the complete picture. When the trajectory reflects typical behavior, the Critic-LLM's diagnosis generalizes and the fix improves the policy. When the trajectory shows an atypical outcome (an unlucky termination, an uncharacteristic failure), the Critic-LLM diagnoses a non-existent problem and destroys behavior that was working in the 19 unseen rollouts. The Critic-LLM cannot distinguish these cases without information about how often the observed failure actually occurs.

\textbf{Failure patterns.}
We categorize AlwaysCritic regressions into four patterns (full descriptions in Appendix~\ref{app:failure-patterns}): \emph{surgical fixes that backfire}, \emph{compulsive revision of near-optimal policies}, \emph{full redesigns on uninformative trajectories}, and \emph{pure stochasticity}. The dominant pattern, compulsive revision, accounts for $12.6\%$ of episodes on Pong and $14.2\%$ on CartPole; pure stochasticity accounts for $<1\%$, confirming noise is not the main driver.

\textbf{Three trajectories and salience bias.}
We tested giving the Critic-LLM the best, worst, and median rollouts (the ThreeTraj variant) to address representativeness. This should, in principle, help distinguish systematic failures from outliers. On Pong, ThreeTraj achieves the worst mean delta per iteration ($+0.16$) despite the largest edits (mean edit distance $8.4$), versus RepTraj's $+0.64$ at distance $6.2$. On CartPole, the same pattern holds: ThreeTraj $+5.6$ at distance $4.6$ vs.\ RepTraj $+16.0$ at distance $4.1$.

To understand why, we coded all 2{,}000 ThreeTraj decisions on CartPole 
and Pong by scanning the Critic-LLM's reasoning text 
(Appendix~\ref{app:salience}). We define a \emph{salience-problem 
regression} as an episode where (i) the Critic-LLM's reasoning explicitly 
references the worst rollout, (ii) the worst rollout's return is strictly 
lower than the median (so the worst is genuinely an outlier), and (iii) 
the revision hurt performance ($\Delta < 0$). On CartPole, 233 of 304 
regressions ($76.6\%$) meet this strict definition. On Pong, where 
rollouts more often agree on a failure (worst $=$ median), ThreeTraj 
instead produces a positive mean delta of $+0.62$ on those episodes.

\begin{quote}
\textbf{Salience bias.} 
\emph{When shown multiple rollouts of a policy, the Critic-LLM systematically overweights the worst trajectory in its diagnosis and edits, even when most rollouts indicate good performance.}
\end{quote}

We observe this directly. In a Pong episode, rollout rewards were $2.0$, $3.0$, $3.0$ (mean $2.95$). The Critic-LLM organized its revision around the worst rollout: ``\textit{From the worst rollout we saw that the policy never moved\ldots this indicates that the do-nothing weights are too dominant.}'' It changed six parameters; reward dropped to $0.15$. Better-performing rollouts existed among the median and best, yet the Critic-LLM anchored on the worst trajectory. This pattern motivates R2PO's evidence design: failure frequency is conveyed through aggregate statistics rather than a worst-case trace.

\textbf{R2PO produces focused, high-quality revisions.}
Table~\ref{tab:revision-summary} reports revision outcomes across the three trajectory-evidence variants; full edit-pattern frequencies are in Appendix~\ref{app:edit-patterns}. R2PO achieves the highest mean improvement per iteration ($+45.3$ on CartPole, $+0.87$ on Pong), substantially outperforming RepTraj ($+16.0$, $+0.64$) and ThreeTraj ($+5.6$, $+0.16$), while making the \emph{smallest} edits (mean edit distance $1.6$ on CartPole, $4.5$ on Pong). It also achieves the highest acceptance rate ($86$--$88\%$) and lowest regression rate ($12$--$14\%$). The pattern breakdown attributes these gains: Near-Optimal Break drops $6\times$ on CartPole and $17\times$ on Pong (Revision rule); Surgical Backfire drops $7\times$ on CartPole and $3.5\times$ on Pong (rollout statistics preventing unrepresentative-trace fixes); productive Fine-Tune edits nearly triple on CartPole ($4.8\% \to 13.4\%$). Consistency across both environments confirms these mechanisms are not environment-specific.

The broader implication is that the main challenge in leveraging trajectories is not \emph{whether} to use them, but \emph{how} to summarize them so the LLM can distinguish systematic failures from rare salient mistakes.

\begin{table}[htbp]
\centering
\caption{Revision summary statistics across 1{,}000 episodes per environment. \colorbox{bestcolor}{\textbf{Best}} per metric is highlighted. R2PO achieves the highest mean improvement while making the smallest edits.}
\label{tab:revision-summary}
\small
\setlength{\tabcolsep}{3pt}
{%
\begin{tabular}{@{}lcccccc@{}}
\toprule
& \multicolumn{3}{c}{CartPole} & \multicolumn{3}{c}{Pong} \\
\cmidrule(lr){2-4} \cmidrule(lr){5-7}
Metric & RepTraj & ThreeTraj & R2PO & RepTraj & ThreeTraj & R2PO \\
\midrule
Mean $\Delta$ (reward/iter) & $+16.0$ & $+5.6$ & \cellcolor{bestcolor}$\mathbf{+45.3}$ & $+0.64$ & $+0.16$ & \cellcolor{bestcolor}$\mathbf{+0.87}$ \\
Mean edit distance & $4.1$ & $4.6$ & \cellcolor{bestcolor}$\mathbf{1.6}$ & $6.2$ & $8.4$ & \cellcolor{bestcolor}$\mathbf{4.5}$ \\
Revision accepted (\%) & 70.7 & 69.6 & \cellcolor{bestcolor}\textbf{86.3} & 71.2 & 67.7 & \cellcolor{bestcolor}\textbf{87.6} \\
Regressions (\%) & 29.3 & 30.4 & \cellcolor{bestcolor}\textbf{13.7} & 28.8 & 32.3 & \cellcolor{bestcolor}\textbf{12.4} \\
\bottomrule
\end{tabular}
}
\end{table}

\textbf{Qualitative examples.}
Figure~\ref{fig:qualitative} shows two representative R2PO episodes: a conservative one-parameter edit guided by statistics, and a well-reasoned minimal edit that destroys the policy and is rejected by selection. Two further examples (full policy rescues on Pong and FrozenLake) and full reasoning traces are in Appendix~\ref{app:qualitative}.

\begin{figure*}[t]
\centering
% Example 1: Conservative Repair
\begin{exbox_q}{Example 1: Conservative Repair (CartPole, R2PO) \hfill \normalfont $490.15$ $\to$ $493.70$ \textcolor{successgreen}{($+3.55$)}}
\small
\textbf{Stats:} Mean 490.15, Min 436, Max 500, Success 19/20 \\[2pt]
\texttt{Initial:~~[6.0, \underline{5.5}, 6.0, 6.0, -1.0, 6.0, -0.5, 6.0, -2.0, -2.0]} \\
\texttt{Revised:~[6.0, \underline{6.0}, 6.0, 6.0, -1.0, 6.0, -0.5, 6.0, -2.0, -2.0]} \\[1pt]
\textbf{Critic-LLM:} \textit{``The average reward (490.15) and success rate (19/20) indicate the policy is already highly effective\ldots I increased only params[1].''} \\[2pt]
\textbf{Takeaway:} Aggregate statistics guide a minimal one-parameter edit. \hfill \textcolor{successgreen}{\textbf{\checkmark~Accepted}}
\end{exbox_q}
% \vspace{1pt}

\begin{exbox_q}{Example 2: Honest Failure (CartPole, R2PO) \hfill \normalfont $436.05$ $\to$ $173.30$ \textcolor{failred}{($-262.75$)}}
\small
\textbf{Stats:} Mean 436.05, Success 15/20, sporadic failures \\[2pt]
\texttt{Initial:~~[6.0, 6.0, 6.0, 6.0, -1.0, 6.0, \underline{-0.5}, 6.0, \underline{-1.8}, -2.0]} \\
\texttt{Revised:~[6.0, 6.0, 6.0, 6.0, -1.0, 6.0, \underline{-0.3}, 6.0, \underline{-1.5}, -2.0]} \\[2pt]
\textbf{Critic-LLM:} \textit{``The failures are sporadic; the median rollout is the same as with the current weights\ldots The adjustments are small and target only the aspects linked to occasional failures.''} \\[1pt]
% \textbf{Takeaway:} A statistically-justified minimal edit destroys the policy. Selection catches the regression. \hfill \textcolor{failred}{\textbf{\texttimes~Original kept}}
\textbf{Takeaway:} Selection catches a minimal edit that destroys the policy. \hfill \textcolor{failred}{\textbf{\texttimes~Original kept}}
\end{exbox_q}
\caption{Two representative R2PO revision episodes. Example 1 shows a conservative one-parameter edit guided by aggregate statistics; Example 2 shows why selection remains essential: well-reasoned minimal edits can still fail. Changed parameters are \underline{underlined}. More examples in Appendix~\ref{app:qualitative}.}
\label{fig:qualitative}
\end{figure*}
\section{Discussion and Conclusion}
\label{sec:discussion}

We introduced R2PO, a two-stage LLM framework that separates global policy proposal (Search-LLM) from local behavior-grounded revision (Critic-LLM), together with a calibrated trajectory evidence package and a selection rule. Our central finding is that rollout trajectories provide a fundamentally richer feedback channel for LLM-based policy optimization than scalar rewards alone, but only when presented carefully: naively passing trajectories to the reviser triggers \emph{salience bias}, in which the LLM overweights vivid worst-case rollouts and over-corrects already-strong policies. R2PO's calibrated evidence package—median trajectory, aggregate statistics, and revision rule—mitigates this bias and yields a more stable and effective optimizer than scalar-only prompting-based search. These results are achieved using a 20B open-weight model, indicating that the benefits of behavior-grounded reflective policy search are not limited to large proprietary systems.

\textbf{Limitations.}
We focus on compact policy classes (tabular and linear); extension to high-dimensional neural policies remains untested. R2PO also uses an environment-dependent threshold $\tau_c$ that currently requires coarse task-specific calibration.  Although Appendix~\ref{app:tau-sensitivity} shows robustness to $\pm 50\%$ variation and competitiveness without this rule, automatic calibration remains unresolved.

\textbf{Future work.}
A first priority is extending R2PO beyond compact controllers. Promising directions include low-dimensional projections of neural network parameters (cf.~\citep{zhou2025prompted}, App.~E) and structured policy classes such as compositional or hierarchical controllers, where parameter substructure could provide the Critic-LLM with semantically meaningful revision targets. A second direction is adaptive calibration of $\tau_c$ from rollout statistics and uncertainty signals, eliminating even the coarse manual setting required today. More broadly, salience bias may affect other settings in which LLMs reason over multiple evidence sources, such as multi-document QA, code review, and multi-turn agents; characterizing this bias beyond policy optimization remains open.

\clearpage
% \section*{References}
\bibliographystyle{plainnat}
\bibliography{references}

%%%%%%%%%%%%%%%%%%%%%%%%%%%%%%%%%%%%%%%%%%%%%%%%%%%%%%%%%%%%
\clearpage
\appendix

\part*{Appendix}
\addcontentsline{toc}{part}{Appendix}
\etocsettocstyle{\section*{Table of Contents}}{}
\etocsetnexttocdepth{subsection}
\localtableofcontents

\clearpage

\section{Significance Tests for Mean-Reward Comparisons}
\label{sec:significance-tests}

To complement the descriptive comparison of reward across methods, this appendix reports significance tests for both \emph{mean reward} and \emph{mean best reward}.
% We report significance tests for \emph{mean reward} since it is the more demanding measure: mean reward reflects the quality of a method throughout training, whereas mean best reward can be driven by a short high-performing peak even if the method is otherwise unstable.

For each environment $e$ and method $m$, let
\[
x^{(e)}_{m,1}, x^{(e)}_{m,2}, \ldots, x^{(e)}_{m,n_m}
\]
denote the \emph{per-run} mean rewards, where each $x^{(e)}_{m,i}$ is the mean reward of run $i$ for method $m$ on environment $e$, averaged across all training iterations, and $n_m = 10$ in our experiments. Thus, the test compares the 10 run-level summary values for R2PO against the 10 run-level summary values for a baseline. The sample mean and (Bessel-corrected) sample variance for method $m$ are

\[
\bar{x}^{(e)}_m = \frac{1}{n_m}\sum_{i=1}^{n_m} x^{(e)}_{m,i},
\qquad
\left(s^{(e)}_m\right)^2 =
\frac{1}{n_m-1}\sum_{i=1}^{n_m}\left(x^{(e)}_{m,i} - \bar{x}^{(e)}_m\right)^2.
\]
The standard deviations reported in the main-paper descriptive tables (e.g., Tables~\ref{tab:main-mean} and~\ref{tab:main-best}) use this same Bessel-corrected sample-variance convention. The reported performance gap is
\[
\Delta^{(e)} = \bar{x}^{(e)}_{\text{R2PO}} - \bar{x}^{(e)}_{\text{baseline}}.
\]

Because the methods were evaluated with independent runs (rather than paired across methods) and exhibited unequal variances in several environments, we use two-sided Welch's t-tests.
\[
t^{(e)} =
\frac{\bar{x}^{(e)}_{\text{R2PO}} - \bar{x}^{(e)}_{\text{baseline}}}
{\sqrt{\frac{\left(s^{(e)}_{\text{R2PO}}\right)^2}{n_{\text{R2PO}}} +
\frac{\left(s^{(e)}_{\text{baseline}}\right)^2}{n_{\text{baseline}}}}},
\]
with Welch--Satterthwaite degrees of freedom
\[
\nu^{(e)} =
\frac{\left(
\frac{\left(s^{(e)}_{\text{R2PO}}\right)^2}{n_{\text{R2PO}}} +
\frac{\left(s^{(e)}_{\text{baseline}}\right)^2}{n_{\text{baseline}}}
\right)^2}
{
\frac{\left(\frac{\left(s^{(e)}_{\text{R2PO}}\right)^2}{n_{\text{R2PO}}}\right)^2}{n_{\text{R2PO}}-1}
+
\frac{\left(\frac{\left(s^{(e)}_{\text{baseline}}\right)^2}{n_{\text{baseline}}}\right)^2}{n_{\text{baseline}}-1}
}.
\]
The corresponding raw two-sided p-value is
\[
p^{(e)}_{\mathrm{raw}} = 2\,\Pr\!\left(T_{\nu^{(e)}} \geq \left|t^{(e)}\right|\right).
\]
Intuitively, a large positive $t$ means that the R2PO runs, viewed as one group, have a higher mean than the baseline runs, viewed as another group, and that this gap is large relative to how much the runs fluctuate within each method. A negative $t$ means the baseline group's mean is higher.

We compare R2PO against \emph{all} core baselines for each environment. That
is, for every environment we test R2PO against ProPS, ProPS+, and the corresponding SB3 baseline. Holm correction is then applied \emph{separately within each baseline family across the 10 environments}; that is, one Holm correction for the 10 R2PO-vs-ProPS comparisons, one for the 10 R2PO-vs-ProPS+ comparisons, and one for the 10 R2PO-vs-SB3 comparisons. If the raw p-values within one such family, sorted in ascending order, are
$p_{(1)}, \ldots, p_{(10)}$, the Holm-adjusted values are computed as
\[
p^{\mathrm{Holm}}_{(k)} =
\max_{j \leq k}\left( (10-j+1)\,p_{(j)} \right),
\]
clipped to $[0,1]$. The raw p-value answers whether a single comparison looks significant on its own; the Holm-adjusted p-value answers whether that comparison remains significant after accounting for the 10-environment family of comparisons against the same baseline type.

Table~\ref{tab:core-significance-mean-all} reports the resulting comparisons for
mean reward. The results are strong. After Holm correction, R2PO significantly
outperforms 28 of the 30 core baseline comparisons. The only non-significant
cases are MountainCarContinuous against SB3 (SAC), where SB3 remains slightly
better in mean reward, and Maze against SB3 (A2C), where the two methods are
effectively tied in mean reward. In contrast, R2PO significantly outperforms ProPS and ProPS+
on all 10 environments.

Table~\ref{tab:core-significance-best-all} reports the corresponding
comparisons for mean best reward. Although only 10 of the 30 mean-best-reward comparisons remain significant after Holm correction, R2PO still matches or exceeds all core baselines descriptively, with positive gaps in most non-significant cases.

These tests reinforce, rather than replace, the descriptive reward tables. The
main value of the significance analysis is to show that R2PO's higher mean
reward is usually not an artifact of run-to-run noise.

% R2PO vs all core baselines (Mean Reward)
\begin{table}[t]
\centering
\small
\caption{Mean-reward significance tests comparing R2PO against all core baselines across the 10 environments. Each comparison uses a two-sided Welch's t-test over 10 independent runs per method; raw p-values are reported together with Holm-corrected p-values, with Holm correction applied separately within each baseline family across the 10 environments. Higher mean reward is highlighted. R2PO significantly outperforms 28 of the 30 core baseline comparisons after correction, with the only non-significant cases being MountainCarContinuous against SB3 (SAC), where SB3 is slightly better, and Maze against SB3 (A2C), where the methods are effectively tied in mean reward.}

\label{tab:core-significance-mean-all}
\resizebox{\textwidth}{!}{%
\begin{tabular}{@{}lrrrrrrr@{}}
\toprule
Environment & Baseline & Base mean & R2PO mean & Gap & Welch $t$ & Raw $p$ & Holm $p$ \\
\midrule
Nim & ProPS & -0.59 & \cellcolor{bestcolor}\textbf{0.61} & 1.20 & 18.122 & 1.5e-08 & 1.1e-07 \\
Nim & ProPS+ & -0.25 & \cellcolor{bestcolor}\textbf{0.61} & 0.86 & 34.411 & 1.5e-12 & 1.5e-11 \\
Nim & SB3 (A2C) & 0.01 & \cellcolor{bestcolor}\textbf{0.61} & 0.60 & 5.958 & 0.0002 & 0.0008 \\
Pong & ProPS & 0.74 & \cellcolor{bestcolor}\textbf{2.51} & 1.77 & 8.435 & 4.2e-06 & 1.7e-05 \\
Pong & ProPS+ & 1.22 & \cellcolor{bestcolor}\textbf{2.51} & 1.28 & 6.588 & 3.5e-05 & 0.0002 \\
Pong & SB3 (PPO) & 1.02 & \cellcolor{bestcolor}\textbf{2.51} & 1.49 & 6.129 & 9.5e-05 & 0.0005 \\
Swimmer & ProPS & 89.22 & \cellcolor{bestcolor}\textbf{260.35} & 171.13 & 8.934 & 9.6e-08 & 5.8e-07 \\
Swimmer & ProPS+ & 162.05 & \cellcolor{bestcolor}\textbf{260.35} & 98.29 & 4.130 & 0.0010 & 0.0031 \\
Swimmer & SB3 (TRPO) & 44.60 & \cellcolor{bestcolor}\textbf{260.35} & 215.75 & 18.549 & 6.4e-09 & 5.1e-08 \\
MountainCarCon & ProPS & -23.45 & \cellcolor{bestcolor}\textbf{81.61} & 105.06 & 7.916 & 2.0e-05 & 6.0e-05 \\
MountainCarCon & ProPS+ & 17.90 & \cellcolor{bestcolor}\textbf{81.61} & 63.71 & 5.324 & 0.0004 & 0.0021 \\
MountainCarCon & SB3 (SAC) & \cellcolor{bestcolor}\textbf{82.33} & 81.61 & -0.72 & -0.393 & 0.7004 & 1.0000 \\
MountainCar & ProPS & -199.31 & \cellcolor{bestcolor}\textbf{-147.84} & 51.47 & 21.891 & 3.2e-10 & 2.6e-09 \\
MountainCar & ProPS+ & -195.81 & \cellcolor{bestcolor}\textbf{-147.84} & 47.97 & 18.897 & 3.7e-11 & 3.3e-10 \\
MountainCar & SB3 (DQN) & -199.99 & \cellcolor{bestcolor}\textbf{-147.84} & 52.15 & 23.196 & 2.4e-09 & 2.2e-08 \\
InvDblPend & ProPS & 79.44 & \cellcolor{bestcolor}\textbf{158.51} & 79.07 & 3.553 & 0.0052 & 0.0052 \\
InvDblPend & ProPS+ & 86.71 & \cellcolor{bestcolor}\textbf{158.51} & 71.80 & 3.213 & 0.0091 & 0.0161 \\
InvDblPend & SB3 (TRPO) & 86.04 & \cellcolor{bestcolor}\textbf{158.51} & 72.47 & 3.350 & 0.0085 & 0.0256 \\
InvPendulum & ProPS & 234.14 & \cellcolor{bestcolor}\textbf{756.08} & 521.93 & 7.190 & 1.1e-06 & 5.7e-06 \\
InvPendulum & ProPS+ & 309.52 & \cellcolor{bestcolor}\textbf{756.08} & 446.55 & 4.834 & 0.0002 & 0.0013 \\
InvPendulum & SB3 (TRPO) & 24.35 & \cellcolor{bestcolor}\textbf{756.08} & 731.73 & 14.977 & 1.1e-07 & 8.0e-07 \\
FrozenLake & ProPS & 0.05 & \cellcolor{bestcolor}\textbf{0.62} & 0.57 & 20.858 & 6.5e-13 & 6.5e-12 \\
FrozenLake & ProPS+ & 0.37 & \cellcolor{bestcolor}\textbf{0.62} & 0.25 & 8.241 & 1.8e-07 & 1.4e-06 \\
FrozenLake & SB3 (TRPO) & 0.02 & \cellcolor{bestcolor}\textbf{0.62} & 0.60 & 24.822 & 5.1e-11 & 5.1e-10 \\
CartPole & ProPS & 258.09 & \cellcolor{bestcolor}\textbf{474.67} & 216.58 & 4.906 & 0.0008 & 0.0015 \\
CartPole & ProPS+ & 253.06 & \cellcolor{bestcolor}\textbf{474.67} & 221.61 & 5.214 & 0.0005 & 0.0021 \\
CartPole & SB3 (TRPO) & 216.92 & \cellcolor{bestcolor}\textbf{474.67} & 257.75 & 12.434 & 1.6e-07 & 9.6e-07 \\
Maze & ProPS & -1.03 & \cellcolor{bestcolor}\textbf{0.83} & 1.86 & 31.357 & 4.2e-12 & 3.8e-11 \\
Maze & ProPS+ & 0.76 & \cellcolor{bestcolor}\textbf{0.83} & 0.07 & 3.001 & 0.0081 & 0.0161 \\
Maze & SB3 (A2C) & 0.83 & \cellcolor{bestcolor}\textbf{0.83} & 0.00 & 0.016 & 0.9878 & 1.0000 \\
\bottomrule
\end{tabular}
}
\end{table}

\begin{table}[t]
\centering
\small
\caption{Best-reward significance tests comparing R2PO against all core baselines across the 10 environments. Each comparison uses a two-sided Welch's t-test over 10 independent runs per method; raw p-values are reported together with Holm-corrected p-values, with Holm correction applied separately within each baseline family across the 10 environments. Higher mean best reward is highlighted. The best-reward differences are less uniformly significant than mean-reward differences: after correction, R2PO significantly outperforms 10 of the 30 core baseline comparisons, and still matches or exceeds all core baselines descriptively, with positive gaps in most non-significant cases.}
% reinforcing that mean best reward is a more optimistic and less stable metric than mean reward.}

\label{tab:core-significance-best-all}
\resizebox{\textwidth}{!}{%
% R2PO vs all core baselines (Mean Best Reward)
\begin{tabular}{@{}lrrrrrrr@{}}
\toprule
Environment & Baseline & Base mean & R2PO mean & Gap & Welch $t$ & Raw $p$ & Holm $p$ \\
\midrule
Nim & ProPS & 0.75 & \cellcolor{bestcolor}\textbf{1.00} & 0.25 & 2.643 & 0.0268 & 0.1338 \\
Nim & ProPS+ & \cellcolor{bestcolor}\textbf{1.00} & \cellcolor{bestcolor}\textbf{1.00} & 0.00 & 0.000 & 1.0000 & 1.0000 \\
Nim & SB3 (A2C) & 0.88 & \cellcolor{bestcolor}\textbf{1.00} & 0.12 & 1.406 & 0.1934 & 0.5803 \\
Pong & ProPS & 2.15 & \cellcolor{bestcolor}\textbf{3.00} & 0.85 & 2.833 & 0.0196 & 0.1177 \\
Pong & ProPS+ & 2.78 & \cellcolor{bestcolor}\textbf{3.00} & 0.22 & 1.687 & 0.1259 & 0.8815 \\
Pong & SB3 (PPO) & 2.33 & \cellcolor{bestcolor}\textbf{3.00} & 0.67 & 2.681 & 0.0251 & 0.1257 \\
Swimmer & ProPS & 208.52 & \cellcolor{bestcolor}\textbf{294.57} & 86.05 & 3.351 & 0.0043 & 0.0299 \\
Swimmer & ProPS+ & 274.86 & \cellcolor{bestcolor}\textbf{294.57} & 19.71 & 1.065 & 0.3011 & 1.0000 \\
Swimmer & SB3 (TRPO) & 67.89 & \cellcolor{bestcolor}\textbf{294.57} & 226.68 & 14.771 & 1.8e-09 & 1.1e-08 \\
MountainCarCon & ProPS & 75.37 & \cellcolor{bestcolor}\textbf{98.75} & 23.38 & 2.418 & 0.0387 & 0.1338 \\
MountainCarCon & ProPS+ & 98.70 & \cellcolor{bestcolor}\textbf{98.75} & 0.05 & 0.237 & 0.8155 & 1.0000 \\
MountainCarCon & SB3 (SAC) & 94.81 & \cellcolor{bestcolor}\textbf{98.75} & 3.94 & 20.521 & 6.6e-14 & 5.2e-13 \\
MountainCar & ProPS & -191.84 & \cellcolor{bestcolor}\textbf{-111.04} & 80.80 & 9.790 & 3.1e-06 & 2.7e-05 \\
MountainCar & ProPS+ & -150.21 & \cellcolor{bestcolor}\textbf{-111.04} & 39.17 & 4.603 & 0.0011 & 0.0115 \\
MountainCar & SB3 (DQN) & -197.47 & \cellcolor{bestcolor}\textbf{-111.04} & 86.44 & 50.957 & 7.6e-21 & 6.9e-20 \\
InvDblPend & ProPS & 112.18 & \cellcolor{bestcolor}\textbf{254.04} & 141.86 & 1.923 & 0.0862 & 0.1724 \\
InvDblPend & ProPS+ & 128.81 & \cellcolor{bestcolor}\textbf{254.04} & 125.23 & 1.659 & 0.1281 & 0.8815 \\
InvDblPend & SB3 (TRPO) & 98.25 & \cellcolor{bestcolor}\textbf{254.04} & 155.79 & 2.120 & 0.0630 & 0.2522 \\
InvPendulum & ProPS & 649.91 & \cellcolor{bestcolor}\textbf{1000.00} & 350.09 & 2.562 & 0.0306 & 0.1338 \\
InvPendulum & ProPS+ & 657.88 & \cellcolor{bestcolor}\textbf{1000.00} & 342.12 & 2.434 & 0.0377 & 0.3394 \\
InvPendulum & SB3 (TRPO) & 28.49 & \cellcolor{bestcolor}\textbf{1000.00} & 971.50 & 4298.503 & 1.0e-29 & 1.0e-28 \\
FrozenLake & ProPS & 0.24 & \cellcolor{bestcolor}\textbf{0.93} & 0.69 & 14.707 & 1.2e-08 & 1.2e-07 \\
FrozenLake & ProPS+ & 0.90 & \cellcolor{bestcolor}\textbf{0.93} & 0.02 & 1.037 & 0.3146 & 1.0000 \\
FrozenLake & SB3 (TRPO) & 0.12 & \cellcolor{bestcolor}\textbf{0.93} & 0.81 & 19.572 & 2.2e-10 & 1.5e-09 \\
CartPole & ProPS & 427.74 & \cellcolor{bestcolor}\textbf{500.00} & 72.26 & 1.598 & 0.1446 & 0.1724 \\
CartPole & ProPS+ & 396.21 & \cellcolor{bestcolor}\textbf{500.00} & 103.79 & 2.115 & 0.0636 & 0.5086 \\
CartPole & SB3 (TRPO) & 490.76 & \cellcolor{bestcolor}\textbf{500.00} & 9.24 & 1.000 & 0.3434 & 0.6869 \\
Maze & ProPS & -0.70 & \cellcolor{bestcolor}\textbf{0.97} & 1.66 & 6.000 & 0.0002 & 0.0016 \\
Maze & ProPS+ & \cellcolor{bestcolor}\textbf{0.97} & \cellcolor{bestcolor}\textbf{0.97} & 0.00 & 0.000 & 1.0000 & 1.0000 \\
Maze & SB3 (A2C) & \cellcolor{bestcolor}\textbf{0.97} & \cellcolor{bestcolor}\textbf{0.97} & 0.00 & 0.000 & 1.0000 & 1.0000 \\
\bottomrule
\end{tabular}
}
\end{table}

\section{Revision threshold sensitivity analysis}
\label{app:tau-sensitivity}

A natural concern about the revision rule (Section~\ref{sec:R2PO-framework}) is whether R2PO's performance depends on careful per-environment tuning of the threshold $\tau_c$. To address this, we conduct a sensitivity analysis on two environments, CartPole and Pong. For each environment, we run R2PO with three threshold settings, all matching the main results' 10-run protocol: (i) the default $\tau_c$ used in the main paper (set close to the known optimum, see Appendix~\ref{app:thresholds}); (ii) $0.5 \cdot \tau_c$, halving the threshold to weaken the revision  rule; and (iii) $\tau_c = -\infty$, which disables the rule entirely.

\begin{table}[h]
\centering
\caption{Revision threshold sensitivity. Each row reports mean and mean best reward across across 10 independent runs for R2PO run with the listed $\tau_c$ value; \emph{Default} matches the value used in the main paper. R2PO's mean reward degrades gracefully as $\tau_c$ is weakened or disabled, but variance increases substantially without the revision rule, particularly on CartPole.}
\label{tab:tau-sensitivity}
\small
\begin{tabular}{@{}lllrrr@{}}
\toprule
Environment & Setting & $\tau_c$ & Mean reward & Mean best reward \\
\midrule
CartPole & default & 480 & $474.67 \pm 16.90$ & $500.00 \pm 0.00$  \\
CartPole & $0.5\tau_c$ & 240 & $446.05 \pm 49.68$ & $489.84 \pm 24.18$  \\
CartPole & disabled & $-\infty$ & $438.36 \pm 63.90$ & $492.49 \pm 23.76$  \\
\midrule
Pong & default & 2.8 & $2.51 \pm 0.21$ & $3.00 \pm 0.00$  \\
Pong & $0.5\tau_c$ & 1.4 & $2.51 \pm 0.48$ & $2.92 \pm 0.24$ \\
Pong & disabled & $-\infty$ & $2.40 \pm 0.59$ & $2.94 \pm 0.21$  \\
\bottomrule
\end{tabular}
\end{table}

\paragraph{Findings.}
R2PO's mean reward is robust across threshold settings, but the revision rule contributes meaningfully to training stability. On CartPole, halving $\tau_c$ reduces mean reward by $\sim$$28$ points and disabling it entirely costs $\sim$$36$ points; standard deviations grow correspondingly from $16.9$ to $49.68$ ($0.5\tau_c$) and $63.9$ (disabled), reflecting more frequent destructive revisions. On Pong, mean reward is essentially preserved across settings, but the standard deviation roughly triples when the revision rule is disabled ($0.21 \to 0.59$), indicating that the rule's main effect on this environment is to suppress occasional regressions rather than to lift mean performance.

Notably, the disabled condition is not a naive baseline: it retains R2PO's two other calibration mechanisms (median-trajectory selection and aggregate rollout statistics). Even without the revision rule, R2PO substantially outperforms the strongest LLM-based baselines (Tables~\ref{tab:main-mean}--\ref{tab:main-best}). This decomposition suggests that the median + statistics package provides the bulk of R2PO's gains by giving the Critic-LLM a representative view of policy behavior, while the rule adds a stability layer that prevents destructive revisions of already-strong policies.

These results suggest that practitioners can apply R2PO with a coarsely chosen $\tau_c$ (e.g., set near the known reward ceiling) without careful per-environment tuning. We discuss adaptive calibration as future work in Section~\ref{sec:discussion}.

\section{Stability analysis: best-vs-mean reward gap}
\label{app:stability}

A method's gap between mean best reward and mean reward across training quantifies its training stability: a small gap indicates that strong policies, once discovered, are reliably maintained; a large gap indicates a method that occasionally finds good policies but loses them across iterations. We compute this gap as $\text{mean best reward} - \text{mean reward}$ for each method-environment pair using the values from Tables~\ref{tab:main-mean} and~\ref{tab:main-best} of the main paper.

\begin{table}[h]
\centering
\caption{Stability gap (mean best reward $-$ mean reward) per method across all 10 environments. Smaller gap indicates more stable training. \colorbox{bestcolor}{\textbf{Smallest gap}} per environment is highlighted. Note that small gaps may also reflect uniformly poor performance (e.g., a method that fails consistently has little to ``lose'' between best and mean); see discussion below.}
\label{tab:stability-gap}
\small
\newsavebox{\stablegapbox}
\sbox{\stablegapbox}{%
\begin{tabular}{@{}lccc@{}}
\toprule
Environment & ProPS & ProPS+ & R2PO \\
\midrule
Nim               & 1.34   & 1.25   & \cellcolor{bestcolor}\textbf{0.39}   \\
Pong              & 1.41   & 1.56   & \cellcolor{bestcolor}\textbf{0.49}   \\
Swimmer           & 119.30 & 112.81 & \cellcolor{bestcolor}\textbf{34.22}  \\
MountainCarCon    & 98.82  & 80.80  & \cellcolor{bestcolor}\textbf{17.14}  \\
MountainCar       & \cellcolor{bestcolor}\textbf{7.47}$^{*}$   & 45.61  & 36.80  \\
InvDblPend        & \cellcolor{bestcolor}\textbf{32.75}$^{*}$  & 42.10  & 95.53  \\
InvPendulum       & 415.77 & 348.35 & \cellcolor{bestcolor}\textbf{243.92} \\
FrozenLake        & \cellcolor{bestcolor}\textbf{0.19}$^{*}$   & 0.54   & 0.31   \\
CartPole          & 169.65 & 143.15 & \cellcolor{bestcolor}\textbf{25.33}  \\
Maze              & 0.34   & 0.21   & \cellcolor{bestcolor}\textbf{0.14}   \\
\bottomrule
\end{tabular}%
}

\usebox{\stablegapbox}

\vspace{2pt}
\parbox{\wd\stablegapbox}{\footnotesize
$^{*}$ Cases where the smallest stability gap reflects uniformly poor performance rather than reliable stability: ProPS reaches near-floor performance on these environments (e.g., MountainCar mean reward $\sim$$-199$ very close to its best $\sim$$-192$ , indicating that the method never finds good policies, rather than stably maintaining them).
}
\end{table}

R2PO exhibits the smallest stability gap on 7 of 10 environments. The three exceptions (MountainCar, InvDblPend, FrozenLake) are not cases where ProPS is genuinely more stable, but rather cases where ProPS reaches near-floor performance and therefore has little spread between its best and mean rewards. On environments where ProPS finds good policies at all (Pong, Swimmer, MountainCarCon, InvPendulum, CartPole), R2PO's gap is consistently smaller. This confirms the qualitative claim in Section~\ref{sec:main-results} that R2PO reliably sustains strong policies once discovered, while prior LLM-based methods exhibit substantially higher run-to-run variability.

\section{Comparison with ProPS at extended training budget}
\label{app:props-comparison}

To contextualize R2PO against the originally published ProPS results, Table~\ref{tab:props-extended} reports our mean best reward (4{,}000 episodes per run, gpt-oss:20b) alongside the ProPS, ProPS+, and SB3 baseline numbers from~\citep{zhou2025prompted} (8{,}000 episodes per run; GPT-4o for ProPS and ProPS+, 10 independent runs). We use the same Gymnasium versions, custom environment specifications, and mean-best-reward metric as~\citep{zhou2025prompted}. The ProPS and ProPS+ numbers in this table differ from the corresponding entries in Tables~\ref{tab:main-mean} and~\ref{tab:main-best} of the main paper, which report our independent matched-budget reproductions (4{,}000 episodes, gpt-oss:20b, 10 independent runs).

Despite training on half the data and using an open-weight 20B-parameter model rather than GPT-4o, R2PO matches or exceeds both prior LLM-based methods and the strongest SB3 baseline on 9 of 10 environments. The exception is Inverted Double Pendulum, where TRPO's reported mean of $3609$ has a standard error of $4000$, indicating one or two outlier runs rather than reliable performance. This result demonstrates that the gains from trajectory-grounded revision are not contingent on access to large proprietary models or extended training budgets.

\begin{table}[h]
\centering
\caption{Mean best reward $\pm$ standard error of the mean over 10 independent runs, matching the convention of~\citep{zhou2025prompted}. Best Baseline, 2nd Best Baseline, ProPS, and ProPS+ numbers are reproduced from~\citep{zhou2025prompted} (8{,}000 episodes per run; GPT-4o for ProPS and ProPS+, 10 runs). R2PO numbers are from our experiments (4{,}000 episodes per run, gpt-oss:20b, 10 runs). \colorbox{bestcolor}{\textbf{Best}} per environment is highlighted across all reported methods.}

\label{tab:props-extended}
\small
\setlength{\tabcolsep}{4pt}
\resizebox{\textwidth}{!}{%
\begin{tabular}{@{}lllccc@{}}
\toprule
Domain & Best Baseline & 2nd Best Baseline & ProPS & ProPS+ & R2PO \\
\midrule
Nim & A2C $0.58 \pm 0.10$ & TRPO $0.50 \pm 0.10$ & $0.33 \pm 0.29$ & $0.97 \pm 0.09$ & \cellcolor{bestcolor}$\mathbf{1.00 \pm 0.00}$ \\
Pong & PPO $2.29 \pm 0.91$ & TRPO $1.36 \pm 1.05$ & $2.80 \pm 0.26$ & $2.99 \pm 0.03$ & \cellcolor{bestcolor}$\mathbf{3.00 \pm 0.00}$ \\
Swimmer & TRPO $52.96 \pm 18.86$ & A2C $39.40 \pm 6.54$ & $218.83 \pm 58.45$ & $227.30 \pm 56.23$ & \cellcolor{bestcolor}$\mathbf{294.57 \pm 13.85}$ \\
MountainCarCon & SAC $86.65 \pm 0.84$ & PPO $78.16 \pm 5.32$ & $87.21 \pm 29.28$ & $89.16 \pm 29.72$ & \cellcolor{bestcolor}$\mathbf{98.75 \pm 0.14}$ \\
MountainCar & DQN $-194.36 \pm 1.47$ & A2C $-200.00 \pm 0.00$ & $-126.11 \pm 21.67$ & $-116.71 \pm 15.20$ & \cellcolor{bestcolor}$\mathbf{-111.04 \pm 1.24}$ \\
InvDblPend & TRPO \cellcolor{bestcolor}$\mathbf{3609.37 \pm 4000.04}$ & PPO $108.60 \pm 4.12$ & $128.17 \pm 24.52$ & $148.39 \pm 48.65$ & $254.04 \pm 73.49$ \\
InvPendulum & TRPO $571.31 \pm 358.88$ & PPO $218.65 \pm 129.31$ & \cellcolor{bestcolor}$\mathbf{1000.00 \pm 0.00}$ & \cellcolor{bestcolor}$\mathbf{1000.00 \pm 0.00}$ & \cellcolor{bestcolor}$\mathbf{1000.00 \pm 0.00}$ \\
FrozenLake & TRPO $0.22 \pm 0.05$ & PPO $0.16 \pm 0.02$ & $0.57 \pm 0.17$ & $0.19 \pm 0.05$ & \cellcolor{bestcolor}$\mathbf{0.93 \pm 0.02}$ \\
CartPole & TRPO $465.34 \pm 62.32$ & PPO $365.86 \pm 73.38$ & $478.27 \pm 65.17$ & \cellcolor{bestcolor}$\mathbf{500.00 \pm 0.00}$ & \cellcolor{bestcolor}$\mathbf{500.00 \pm 0.00}$ \\
Maze & A2C \cellcolor{bestcolor}$\mathbf{0.97 \pm 0.00}$ & TRPO \cellcolor{bestcolor}$\mathbf{0.97 \pm 0.00}$ & $0.55 \pm 0.83$ & \cellcolor{bestcolor}$\mathbf{0.97 \pm 0.00}$ & \cellcolor{bestcolor}$\mathbf{0.97 \pm 0.00}$ \\
\bottomrule
\end{tabular}
}
\end{table}

\section{R2PO algorithm}
\label{app:algorithm}

Algorithm~\ref{alg:R2PO} provides the full R2PO optimization loop. At each iteration, the Search-LLM proposes an initial parameter vector from the reward-only replay history, the environment evaluates it over $K$ rollouts, and the resulting trajectories are summarized into a calibrated evidence package (median trajectory, aggregate statistics, and a revision rule controlled by $\tau_c$). The Critic-LLM uses this evidence to propose a revised parameter vector, which is re-evaluated; the keep-best selection rule retains whichever candidate achieves the higher mean reward and appends it to the replay buffer.

\begin{algorithm}[h]
\caption{R2PO}
\label{alg:R2PO}
\begin{algorithmic}[1]
\Require environment \textsc{env}, environment description \texttt{env\_desc}, rollouts per evaluation $K$, iterations $T$, Search-LLM prompt $P_1$, Critic-LLM prompt $P_2$, revision threshold $\tau_c$
\State $\textsc{replay\_buffer} \gets \emptyset$
\For{$t = 1, \ldots, T$}
    \State $H \gets \textsc{FormatHistory}(\textsc{replay\_buffer})$ \Comment{reward-only: $\{(\theta_i, \bar{R}_i)\}$}
    \State $\theta_{\text{init}} \gets \textsc{Search-LLM}(P_1, H)$ \Comment{Stage 1: proposal}
    \State $\bar{R}_{\text{init}}, \mathcal{T}_{\text{init}} \gets \textsc{Eval}(\textsc{env}, \theta_{\text{init}}, K)$
    \State $\tau^* \gets \textsc{SelectMedian}(\mathcal{T}_{\text{init}})$
    \State stats $\gets \textsc{AggregateStats}(\mathcal{T}_{\text{init}})$
    \State evidence $\gets \textsc{BuildContext}(\theta_{\text{init}}, \bar{R}_{\text{init}}, \tau^*, \text{stats}, H, \tau_c, \texttt{env\_desc})$
    \State $\theta_{\text{rev}} \gets \textsc{Critic-LLM}(P_2, \text{evidence})$ \Comment{Stage 2: reflection}
    \State $\bar{R}_{\text{rev}} \gets \textsc{Eval}(\textsc{env}, \theta_{\text{rev}}, K)$
    \If{$\bar{R}_{\text{rev}} \geq \bar{R}_{\text{init}}$} \Comment{selection}
        \State $\textsc{replay\_buffer}.\textsc{add}(\theta_{\text{rev}}, \bar{R}_{\text{rev}})$
    \Else
        \State $\textsc{replay\_buffer}.\textsc{add}(\theta_{\text{init}}, \bar{R}_{\text{init}})$
    \EndIf
\EndFor
\State \Return \textsc{replay\_buffer}
\end{algorithmic}
\end{algorithm}

\section{Revision threshold values}
\label{app:thresholds}
 
Table~\ref{tab:thresholds} reports the revision threshold $\tau_c$ used by R2PO for each environment. The threshold is set close to the strongest reward empirically achievable in our setup, which equals the theoretical optimum on tasks with bounded returns and is set lower on tasks where the formal ceiling is far above what any method reaches under our budget.

\begin{table}[ht]
\centering
\caption{Revision threshold $\tau_c$ for each environment. Set close to the strongest reward empirically achievable in our setup; equals the theoretical optimum on bounded-return tasks (Nim, Pong, MountainCar, FrozenLake, CartPole, Maze) and falls below it on tasks with very high theoretical ceilings (e.g., InvDblPend, where $\tau_c = 450$ vs. a formal optimum of $10{,}000$).}
\label{tab:thresholds}
\small
\begin{tabular}{@{}lrr@{}}
\toprule
Environment & $\tau_c$ & Optimal reward \\
\midrule
Nim & 0.95 & 1.00$^{*}$ \\
Pong & 2.80 & 3.00$^{*}$ \\
Swimmer & 230 & 250$^{*}$ \\
MountainCarCon & 97 & 100$^{\S}$ \\
MountainCar & $-120$ & $-120^{*}$ \\
InvDblPend & 450 & 10000$^{\S}$ \\
InvPendulum & 950 & 1000$^{\S}$ \\
FrozenLake & 0.85 & 1.00$^{\ddagger}$ \\
CartPole & 480 & 500$^{\dagger}$ \\
Maze & 0.90 & 0.97$^{*}$ \\
\bottomrule
\multicolumn{3}{@{}p{\linewidth}@{}}{\footnotesize
$^{\dagger}$ official Gymnasium reward threshold (and, for CartPole-v1, also the effective episode maximum);
$^{\ddagger}$ episode maximum return;
$^{\S}$ theoretical reward ceiling implied by the reward function and horizon, not an official Gymnasium threshold;
$^{*}$ paper reference reward used to set $\tau_c$; official Gymnasium thresholds, where available, may differ.}
\end{tabular}
\end{table}

\section{Full learning curves}
\label{app:learning-curves}

\begin{figure*}[h]
\centering
\includegraphics[width=\textwidth]{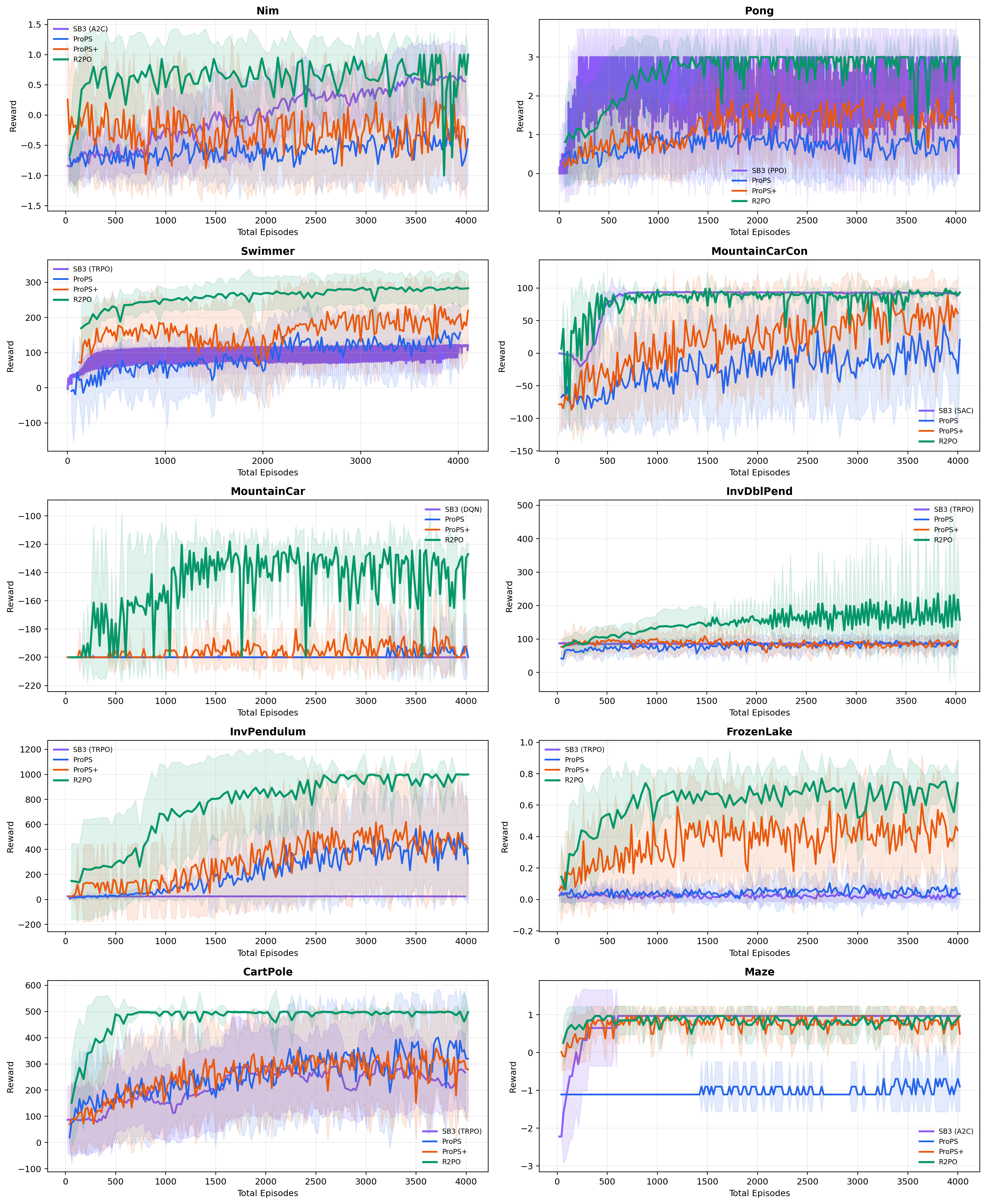}
\caption{Learning curves (mean $\pm$ standard deviation over 10 independent runs) for R2PO, ProPS, ProPS+, and the best SB3 baseline across all ten environments.}
\label{fig:learning-curves-full}
\end{figure*}

Figure~\ref{fig:learning-curves-full} shows learning curves (mean $\pm$ standard deviation over 10 independent runs) for R2PO, ProPS, ProPS+, and the best SB3 baseline across all ten environments. R2PO generally reaches strong performance earlier and maintains it more consistently than ProPS and ProPS+, which often exhibit higher variance and less stable training. 

\section{Mean best reward for ablation variants}
\label{app:variants-best}

Table~\ref{tab:variants-best} reports mean best reward across 10 independent runs for R2PO and all ablation variants. The pattern matches mean reward (Table~\ref{tab:variants} in the main text): R2PO matches or exceeds the strongest variant on all environments.

% ── Variants Table: Mean Best Reward ──
\begin{table}[t]
\centering
\caption{Mean best reward ($\pm$ standard deviation) across 10 independent runs for R2PO and the ablation variants, where each run's best is taken across training iterations. \colorbox{bestcolor}{\textbf{Best}} per environment is highlighted. BestBaseline denotes the best among ProPS, ProPS+, and all SB3 algorithms. R2PO matches or exceeds the strongest variant on all environments.}
\label{tab:variants-best}
\small
\setlength{\tabcolsep}{2pt}
\resizebox{\textwidth}{!}{%
\begin{tabular}{@{}lcccccccc@{}}
\toprule
Environment & BestBaseline & PureSearch & ActorSecondPass & CriticOnly & AlwaysCritic & RepTraj & ThreeTraj & R2PO \\
\midrule
Nim & \cellcolor{bestcolor}$\mathbf{1.00 \pm 0.00}$ {\tiny(ProPS+)} & $0.70 \pm 0.29$ & $0.45 \pm 0.46$ & \cellcolor{bestcolor}$\mathbf{1.00 \pm 0.00}$ & \cellcolor{bestcolor}$\mathbf{1.00 \pm 0.00}$ & \cellcolor{bestcolor}$\mathbf{1.00 \pm 0.00}$ & \cellcolor{bestcolor}$\mathbf{1.00 \pm 0.00}$ & \cellcolor{bestcolor}$\mathbf{1.00 \pm 0.00}$ \\
Pong & $2.78 \pm 0.41$ {\tiny(ProPS+)} & $2.31 \pm 0.62$ & $2.18 \pm 0.63$ & \cellcolor{bestcolor}$\mathbf{3.00 \pm 0.00}$ & $2.83 \pm 0.54$ & $3.00 \pm 0.02$ & $2.74 \pm 0.41$ & \cellcolor{bestcolor}$\mathbf{3.00 \pm 0.00}$ \\
Swimmer & $274.86 \pm 38.81$ {\tiny(ProPS+)} & $219.56 \pm 59.49$ & $175.38 \pm 83.16$ & $210.37 \pm 77.23$ & $264.22 \pm 72.99$ & $257.80 \pm 63.12$ & $178.71 \pm 92.87$ & \cellcolor{bestcolor}$\mathbf{294.57 \pm 43.79}$ \\
MountainCarCon & $98.70 \pm 0.50$ {\tiny(ProPS+)} & $95.31 \pm 7.96$ & $75.40 \pm 40.01$ & $83.52 \pm 29.86$ & $70.55 \pm 42.99$ & $46.70 \pm 49.24$ & $68.90 \pm 47.57$ & \cellcolor{bestcolor}$\mathbf{98.75 \pm 0.44}$ \\
MountainCar & $-150.21 \pm 26.62$ {\tiny(ProPS+)} & $-170.22 \pm 41.59$ & $-191.71 \pm 26.22$ & $-114.68 \pm 4.04$ & $-117.16 \pm 7.37$ & $-128.69 \pm 32.99$ & $-118.60 \pm 16.68$ & \cellcolor{bestcolor}$\mathbf{-111.04 \pm 3.92}$ \\
InvDblPend & $128.81 \pm 54.51$ {\tiny(ProPS+)} & $121.84 \pm 39.64$ & $112.29 \pm 16.56$ & $232.67 \pm 106.26$ & $207.42 \pm 86.19$ & $211.30 \pm 91.05$ & $124.09 \pm 38.11$ & \cellcolor{bestcolor}$\mathbf{254.04 \pm 232.39}$ \\
InvPendulum & $657.88 \pm 444.42$ {\tiny(ProPS+)} & $720.46 \pm 451.28$ & $810.05 \pm 400.62$ & $365.45 \pm 380.21$ & $287.69 \pm 273.42$ & $815.09 \pm 298.26$ & $357.20 \pm 445.93$ & \cellcolor{bestcolor}$\mathbf{1000.00 \pm 0.00}$ \\
FrozenLake & $0.90 \pm 0.04$ {\tiny(ProPS+)} & $0.29 \pm 0.17$ & $0.24 \pm 0.08$ & $0.90 \pm 0.04$ & $0.88 \pm 0.06$ & $0.91 \pm 0.03$ & $0.91 \pm 0.04$ & \cellcolor{bestcolor}$\mathbf{0.93 \pm 0.05}$ \\
CartPole & $490.76 \pm 29.22$ {\tiny(TRPO)} & $440.09 \pm 129.94$ & $461.59 \pm 99.44$ & $498.62 \pm 3.45$ & $498.54 \pm 4.62$ & $487.74 \pm 38.79$ & \cellcolor{bestcolor}$\mathbf{500.00 \pm 0.00}$ & \cellcolor{bestcolor}$\mathbf{500.00 \pm 0.00}$ \\
Maze & \cellcolor{bestcolor}$\mathbf{0.97 \pm 0.00}$ {\tiny(ProPS+)} & $-0.90 \pm 0.66$ & $-0.90 \pm 0.66$ & \cellcolor{bestcolor}$\mathbf{0.97 \pm 0.00}$ & \cellcolor{bestcolor}$\mathbf{0.97 \pm 0.00}$ & \cellcolor{bestcolor}$\mathbf{0.97 \pm 0.00}$ & \cellcolor{bestcolor}$\mathbf{0.97 \pm 0.00}$ & \cellcolor{bestcolor}$\mathbf{0.97 \pm 0.00}$ \\
\bottomrule
\end{tabular}

}
\end{table}

\section{Salience analysis details}
\label{app:salience}

To quantify the salience bias observed in the ThreeTraj variant, we 
systematically analyzed all 2{,}000 ThreeTraj episodes across CartPole 
and Pong (1{,}000 each, 10 independent runs per environment). For each 
episode, we parsed Critic-LLM's reasoning text from the logged output, 
checking whether the worst rollout was explicitly mentioned, and computed 
the reward delta $\Delta = R_{\text{revised}} - R_{\text{initial}}$.

\paragraph{Operational definition.} An episode is classified as a 
\emph{salience-problem regression} if all of the following hold:
\begin{enumerate}[nosep,leftmargin=*]
\item Critic-LLM's reasoning explicitly references the worst rollout.
\item The worst rollout's return is strictly lower than the median 
(\emph{strict} criterion), so the worst is genuinely an outlier rather 
than the modal outcome.
\item The revision hurt performance: $\Delta < 0$.
\end{enumerate}
We also report a \emph{permissive} variant that relaxes condition (2) 
to require only that the worst be no greater than the median and 
strictly less than the best. The strict criterion isolates 
outlier-driven misdiagnosis; the permissive criterion additionally 
counts cases where the worst trajectory reflects a $\geq 50\%$ failure 
mode but is still cited by the Critic-LLM as the central evidence for 
revision.

\paragraph{Results.}

\begin{table}[h]
\centering
\caption{Salience analysis across 2{,}000 ThreeTraj episodes (1{,}000 each 
on CartPole and Pong, 10 runs per environment). Under the strict criterion, 
salience-problem regressions account for 76.6\% of CartPole regressions and 
31.3\% of Pong regressions. The gap reflects task structure: on CartPole, 
all 304 regressions have worst $<$ median, so strict and permissive criteria 
coincide; on Pong, $61.5\%$ of regressions have worst $=$ median, where the 
worst trajectory is a typical failure mode rather than an outlier.}
\label{tab:salience}
\small
\begin{tabular}{@{}lrr@{}}
\toprule
Metric & CartPole & Pong \\
\midrule
Total episodes & 1{,}000 & 1{,}000 \\
Total regressions & 304 (30.4\%) & 323 (32.3\%) \\
\midrule
\multicolumn{3}{@{}l}{\emph{Salience-problem regressions (\% of regressions)}} \\
Strict (worst $<$ median) & 233 (76.6\%) & 101 (31.3\%) \\
Permissive (worst $\leq$ median $<$ best) & 233 (76.6\%) & 261 (80.8\%) \\
\bottomrule
\end{tabular}
\end{table}

The strict salience proxy yields different rates on CartPole (76.6\%) 
and Pong (31.3\%) because the two environments differ in how often the 
worst rollout is genuinely an outlier. On CartPole, all 304 regression 
episodes have worst $<$ median, so the strict and permissive criteria 
coincide: in 76.6\% of regression cases, the Critic-LLM cited an outlier 
worst-case rollout and produced a harmful revision. On Pong, $61.5\%$ of 
regression episodes have worst $=$ median, where the worst trajectory is 
also typical and Critic-LLM's reference to it is closer to legitimate 
diagnosis; the strict criterion correctly excludes these cases, leaving 
31.3\% as true outlier-driven regressions.

The Pong rollout-agreement breakdown (Table~\ref{tab:salience-pong-breakdown}) 
further supports the diagnosis: when failures are confirmed across 
rollouts (worst $=$ median), ThreeTraj produces a positive mean delta 
($+0.62$), indicating that trajectory-grounded diagnosis works well when 
the worst trajectory is representative. When rollouts suggest the policy 
is already strong (preservation-favored), the mean delta is $-0.35$, 
indicating that Critic-LLM tends to damage policies it should have 
preserved. Together, these results support the design decision in R2PO 
to exclude the worst-case trajectory from Critic-LLM's context and 
instead provide failure frequency through aggregate rollout statistics.

\begin{table}[h]
\centering
\caption{Pong rollout-agreement breakdown of the 1{,}000 ThreeTraj 
episodes. The mean revision delta varies sharply with whether the worst 
rollout is representative of typical policy behavior.}
\label{tab:salience-pong-breakdown}
\small
\begin{tabular}{@{}lrr@{}}
\toprule
Rollout-agreement structure & Episodes & Mean $\Delta$ \\
\midrule
Failure-confirmed (worst $=$ median $<$ best) & 491 & $+0.62$ \\
Preservation-favored (median $=$ best $>$ worst) & 382 & $-0.35$ \\
All-diverse (worst $<$ median $<$ best) & 127 & $-0.09$ \\
\bottomrule
\end{tabular}
\end{table}

\paragraph{Qualitative examples of salience-driven regressions.}

We provide three representative ThreeTraj episodes where Critic-LLM anchors on the worst rollout despite stronger evidence from the median and/or best rollouts. In all three cases, the initial policy is already strong on at least two of the three rollouts, Critic-LLM diagnoses the failure primarily through the worst rollout, and the revision reduces the mean reward. Traces are copied verbatim from logged output files, with key evidence phrases \hl{highlighted}.

\subsection*{Salience Example 1: Full Reasoning (Pong, run 10, episode 92)}
 
\textbf{Rollouts:} worst $= 2.0$, median $= 3.0$, best $= 3.0$ \quad \textbf{Initial:} $2.95$ \quad \textbf{Revised:} $0.15$ \quad $\Delta = -2.80$
 
\noindent\textbf{Critic-LLM output:}
 
{\small\ttfamily params[0]: 2.4, params[1]: -3.0, params[2]: 0.0, params[3]: 1.0, params[4]: -1.0, params[5]: 0.4, params[6]: -1.5, params[7]: 1.0, params[8]: 0.0, params[9]: 0.8, params[10]: 0.0, params[11]: 0.8, params[12]: 0.0, params[13]: 0.8, params[14]: 0.0, params[15]: -1.0, params[16]: 1.0, params[17]: -1.0}
 
\begin{quote}\small\ttfamily
I only adjusted the weights that directly influence when the paddle chooses to move up or down versus staying still. \hl{From the worst rollout we saw that the policy never moved} \hl{(action 2 every step)}, even as the ball's y-coordinate changed steadily. \hl{This indicates that the ``do-nothing'' weights are too dominant}, especially the positive weight on ball x (param 5 = 0.8) and the negative weight on ball vx (param 11 = $-0.8$). In the median and best rollouts the paddle sometimes moves at the end of the episode, but the long-term reward only reaches 3.0, so a modest shift toward proactive motion should raise the average reward without over-reacting.
 
Specific changes: param 5 reduced from 0.8 to 0.4; param 6 increased from $-2.8$ to $-1.5$; param 7 reduced from 2.8 to 1.0; param 10 set to 0.0; param 11 changed sign from $-0.8$ to $+0.8$; param 17 moved from 0.0 to $-1.0$.
\end{quote}
 
\textbf{Salience evidence:} Critic-LLM organized all six parameter changes around the worst rollout's ``never moved'' behavior, despite the median and best rollouts already achieving the maximum reward of 3.0. The result was a catastrophic drop from 2.95 to 0.15.
 
\subsection*{Salience Example 2: Full Reasoning (Pong, run 8, episode 9)}
 
\textbf{Rollouts:} worst $= 0.0$, median $= 3.0$, best $= 3.0$ \quad \textbf{Initial:} $2.75$ \quad \textbf{Revised:} $0.15$ \quad $\Delta = -2.60$
 
\noindent\textbf{Critic-LLM output:}
 
{\small\ttfamily params[0]: 1.5, params[1]: -1.5, params[2]: -0.5, params[3]: 0.5, params[4]: -0.5, params[5]: 0.5, params[6]: -1.0, params[7]: 1.0, params[8]: -1.0, params[9]: 1.0, params[10]: -1.0, params[11]: 1.0, params[12]: -0.5, params[13]: 0.5, params[14]: -0.5, params[15]: 0.5, params[16]: -0.5, params[17]: 0.5}
 
\begin{quote}\small\ttfamily
I kept the bias terms moderate to prevent the action scores from drifting too high or low. The weight pattern now favors the paddle to move up when the ball's y-coordinate is below the paddle's y, and to move down when it is above. This mirrors the behavior seen in the successful median/best rollouts.
 
\hl{In the worst rollout, the ball was moving upward (vy = 3.31)} \hl{but the policy's current weights caused it to over-react} to the ball's x and y independently, making the paddle lag behind the ball's trajectory. The new weight set should reduce that lag.
 
The training trial with parameters near 2.0 (trial 26) achieved a high mean\_reward of 2.90; the new parameters preserve that general structure while slightly amplifying the paddle\_y influence. This focused adjustment should bring the average reward closer to the target of $\sim$3 \hl{while avoiding over-fitting to the rare failure case} \hl{seen only in rollout 17}.
\end{quote}
 
\textbf{Salience evidence:} Critic-LLM acknowledges the failure was ``rare'' and ``seen only in rollout 17,'' yet restructures the entire weight pattern to address it. Two rollouts already achieved the maximum reward (3.0), but the revision was driven by the single zero-reward rollout.
 
\subsection*{Salience Example 3: Full Reasoning (CartPole, run 5, episode 67)}
 
\textbf{Rollouts:} worst $= 446$, median $= 500$, best $= 500$ \quad \textbf{Initial:} $497.30$ \quad \textbf{Revised:} $9.55$ \quad $\Delta = -487.75$

\noindent\textbf{Critic-LLM output:}
 
{\small\ttfamily params[0]: 1.5, params[1]: -0.6, params[2]: -4.8, params[3]: 1.0, params[4]: -1.1, params[5]: -1.2, params[6]: 2.8, params[7]: 2.5, params[8]: 2.6, params[9]: 2.4}
 
\begin{quote}\small\ttfamily
\hl{The worst rollout repeatedly toggled actions} \hl{when the pole angle was only a few degrees off-balance}, leading to a reward drop to $\sim$446. In the best and median trajectories the agent consistently chose the action that kept the pole upright even for small angle deviations.
 
I reduced the magnitude of both pole-angle weights: params[4] from $-1.3$ to $-1.1$, and params[5] from $-1.4$ to $-1.2$. This makes the decision boundary for the angle less steep, allowing the policy to keep the same action longer when the pole is close to upright.
 
Additionally \hl{the worst rollout tended to ``over-react''} \hl{after large cart velocity spikes}. Lowering the bias of the right-push column (params[9]) from 2.6 to 2.4 forces the agent to be slightly more cautious.
 
A slightly higher weight on the cart-position component for the left-push column (params[0] to 1.5) and a modest increase in the cart-angle weight for the right-push column (params[7] to 2.5) help balance the left and right pushes, yielding smoother trajectories.
\end{quote}
 
\textbf{Salience evidence:} Critic-LLM changed five parameters, all motivated by the worst rollout's behavior (``toggled actions,'' ``over-react''). The policy was nearly perfect (mean 497.3, median and best both 500). The result was the most catastrophic regression in our analysis: from 497.30 to 9.55.

\subsection{Failure pattern definitions}
\label{app:failure-patterns}

We categorize AlwaysCritic regressions into four patterns:

\begin{itemize}[leftmargin=*]
\item \emph{Surgical fixes that backfire}: a 1--3 parameter edit is locally correct for the observed trace but disrupts behavior in other rollouts.
\item \emph{Compulsive revision of near-optimal policies}: the Critic-LLM perturbs a high-reward policy citing risks that have not manifested ($12.6\%$ of episodes on Pong, $14.2\%$ on CartPole).
\item \emph{Full redesigns on uninformative trajectories}: near-zero policies whose short rollout traces prompt the Critic-LLM to redesign the policy from scratch.
\item \emph{Pure stochasticity}: identical revisions yield different rewards due to rollout variance ($<1\%$ of episodes).
\end{itemize}

\subsection{Edit pattern frequencies}
\label{app:edit-patterns}

Table~\ref{tab:patterns} reports the frequency of distinct edit patterns across the three trajectory-evidence variants. Regression patterns (Near-Optimal Break, Surgical Backfire) capture cases where the Critic-LLM's revision hurts performance; improvement patterns (Pinpoint Fix, Fine-Tune) capture cases where it helps. Pattern definitions are in Appendix~\ref{app:failure-patterns}.

\begin{table}[h]
\centering
\caption{Edit pattern frequencies (\% of 1{,}000 episodes). \colorbox{bestcolor}{\textbf{Best}} per pattern is highlighted. R2PO sharply reduces destructive patterns while increasing productive ones on both environments.}

\label{tab:patterns}
\small
\setlength{\tabcolsep}{3pt}
{%
\begin{tabular}{@{}lcccccc@{}}
\toprule
& \multicolumn{3}{c}{CartPole} & \multicolumn{3}{c}{Pong} \\
\cmidrule(lr){2-4} \cmidrule(lr){5-7}
Pattern & RepTraj & ThreeTraj & R2PO & RepTraj & ThreeTraj & R2PO \\
\midrule
\multicolumn{7}{@{}l}{\emph{Regression patterns (lower is better)}} \\
\quad Near-Optimal Break & 13.3 & 18.6 & \cellcolor{bestcolor}\textbf{2.1} & 10.5 & 6.0 & \cellcolor{bestcolor}\textbf{0.6} \\
\quad Surgical Backfire & 6.0 & 1.7 & \cellcolor{bestcolor}\textbf{0.8} & 2.8 & 1.2 & \cellcolor{bestcolor}\textbf{0.8} \\
\midrule
\multicolumn{7}{@{}l}{\emph{Improvement patterns (higher is better)}} \\
\quad Pinpoint Fix & 4.1 & \cellcolor{bestcolor}\textbf{6.8} & 6.2 & 3.5 & 1.6 & \cellcolor{bestcolor}\textbf{3.7} \\
\quad Fine-Tune & 4.8 & 8.1 & \cellcolor{bestcolor}\textbf{13.4} & 0.1 & 0.0 & \cellcolor{bestcolor}\textbf{1.3} \\
\bottomrule
\end{tabular}
}
\end{table}

\section{Full qualitative reasoning traces}
\label{app:qualitative}

We provide complete Critic-LLM reasoning output for the two examples in Figure~\ref{fig:qualitative}, plus three additional revision episodes: two full policy rescues (Pong: $0.05 \to 3.00$ and FrozenLake: $0.00 \to 0.90$) and a loop-diagnosis case. Traces are copied verbatim from logged output files, with key evidence phrases \hl{highlighted}.
 
\subsection*{Conservative Repair (CartPole, run 10, episode 26)}
 
\textbf{Initial reward:} 490.15 \quad \textbf{Revised reward:} 493.70 \quad $\Delta = +3.55$
 
\noindent\textbf{Stats given to Critic-LLM:}
\begin{quote}\small\ttfamily
Reward: mean=490.15, min=303.00, max=500.00\\
Episode length: mean=490.1, min=303, max=500\\
Success rate: 19/20 rollouts reached reward=500.00\\
Failure rate: 1/20 rollouts finished below reward=500.00\\
Median rollout (rollout 0, reward=500.0000, length=500)
\end{quote}
 
\noindent\textbf{Critic-LLM reasoning:}
\begin{quote}\small\ttfamily
params[0]: 6, params[1]: 6, params[2]: 6, params[3]: 6, params[4]: -1, params[5]: 6, params[6]: -0.5, params[7]: 6, params[8]: -2, params[9]: -2
 
The one-step failure at 303 was caused by a slightly weaker push to the right (action 1) when the pole angle was slightly positive and the cart velocity was positive. In the median trajectory this situation never caused an issue; the policy performed robustly for all 500 steps. \hl{The average reward (490.15) and success rate (19/20)} \hl{indicate that the policy is already highly effective}, with only one occasional sub-maximal episode.
 
\hl{To eliminate this outlier without risk of over-fitting, I increased only} \hl{params[1]}, which controls the contribution of cart velocity to the left-push action, from 5.5 to the maximum of 6.0. This mirrors the successful configuration from trial 34, which achieved perfect 500-step episodes across all rollouts. \hl{All other parameters are kept identical to preserve the observed strong performance.}
\end{quote}
 
\subsection*{Full Rescue (Pong, run 10, episode 86)}
 
\textbf{Initial reward:} 0.05 \quad \textbf{Revised reward:} 3.00 \quad $\Delta = +2.95$
 
\noindent\textbf{Stats given to Critic-LLM:}
\begin{quote}\small\ttfamily
Reward: mean=0.05, min=0.00, max=1.00\\
Episode length: mean=180.0, min=81, max=541\\
Success rate: 1/20 rollouts reached reward=1.00\\
Failure rate: 19/20 rollouts finished below reward=1.00\\
Median rollout (rollout 0, reward=0.0000, length=233)
\end{quote}
 
\noindent\textbf{Critic-LLM reasoning:}
\begin{quote}\small\ttfamily
params[0]: 0.9, params[1]: -0.9, params[2]: 0.0, params[3]: 0.0, params[4]: 0.0, params[5]: 0.0, params[6]: -0.9, params[7]: 0.9, params[8]: 0.0, params[9]: 0.0, params[10]: 0.0, params[11]: 0.0, params[12]: 0.0, params[13]: 0.0, params[14]: 0.0, params[15]: 0.0, params[16]: 0.0, params[17]: 0.0
 
\hl{The current policy almost never moves the paddle} because the weight on ball\_x (params[3]) pulls action 0 down as the ball travels rightward, while the action-2 "do nothing" score is always higher than the up/down scores. \hl{The median rollout shows the paddle staying at y = 300} \hl{for most of the episode} even though the ball's y grows steadily from 300 to \textasciitilde500, so the relative vertical distance is never used to drive a move.
 
\hl{The aggregate statistics} (many successful trials) \hl{indicate that the best performing policies depend only on} \hl{the paddle's y position and the ball's y position}, with the following relation:
 
Action 0 (up) score = pad\_y - ball\_y\\
Action 1 (down) score = -pad\_y + ball\_y\\
Action 2 (stay) score = 0
 
This simple comparison perfectly tracks the ball's vertical motion and yields a reward of +1 in every episode.
 
The new proposal keeps this structure but chooses a scaling that has not been tried before (0.9 instead of the exact 1.0 or 1.2 used in earlier trials). Because both action 0 and action 1 are scaled equally, the relative ordering of their scores is unchanged, ensuring identical behaviour while satisfying the rule against reproducing existing parameter sets.
\end{quote}
 
\subsection*{Full Rescue (FrozenLake, run 10, episode 10)}
 
\textbf{Initial reward:} 0.00 \quad \textbf{Revised reward:} 0.90 \quad $\Delta = +0.90$
 
\noindent\textbf{Stats given to Critic-LLM:}
\begin{quote}\small\ttfamily
Reward: mean=0.00, min=0.00, max=0.00\\
Episode length: mean=100.0, min=100, max=100\\
Success rate: 0/20 rollouts reached reward=1.00\\
Failure rate: 20/20 rollouts finished below reward=1.00\\
Median rollout (rollout 0, reward=0.0000, length=100, outcome=reached the rollout cap)
\end{quote}
 
\noindent\textbf{Critic-LLM reasoning:}
\begin{quote}\small\ttfamily
params[0]: 0, params[1]: 3, params[2]: 0, params[3]: 3, params[4]: 0, params[5]: 0, params[6]: 0, params[7]: 1, params[8]: 3, params[9]: 1, params[10]: 0, params[11]: 2, params[12]: 0, params[13]: 2, params[14]: 1, params[15]: 3

\hl{I revisited the "median rollout" and the average statistics} \hl{to pinpoint where the all-3 policy failed.}
* The agent kept attempting the "left/up/right" mix (action 3), \hl{causing it to repeatedly bounce in place} \hl{or steer into holes with probability 1/3.}
* States near the bottom and right edge (7, 11, 14) were exposed to a high chance of stepping into a hole when using action 0 or 3.
I therefore kept the successful choices from Trial 2 (e.g., 0 at state 0, 3 at state 1, etc.) but made two targeted changes that reduce hole-risk:

1. \textbf{State 7} - originally action 3 had a 1/3 chance of moving up into hole 3. Replaced by action 1, which only risks moving left to safe state 6 or staying, eliminating the up-hole accidental step.
2. \textbf{State 11} - remaining in column 3, action 0 risked falling left into hole 10. Switched to action 2, which keeps the agent within the goal corridor (up to 7 or down to 15) and never reaches the hole directly.

All other actions retain the values that previously yielded high reward (0.8 in Trial 2). \hl{This modest, focused adjustment is expected} \hl{ to increase the success rate} \hl{toward the target $\approx 1.0$} while avoiding the stochastic pitfalls that caused the all-3 policy to fail.
\end{quote}
 
\subsection*{Honest Failure (CartPole, run 10, episode 16)}
 
\textbf{Initial reward:} 436.05 \quad \textbf{Revised reward:} 173.30 \quad $\Delta = -262.75$
 
\noindent\textbf{Stats given to Critic-LLM:}
\begin{quote}\small\ttfamily
Reward: mean=436.05, min=204.00, max=500.00\\
Episode length: mean=436.1, min=204, max=500\\
Success rate: 15/20 rollouts reached reward=500.00\\
Failure rate: 5/20 rollouts finished below reward=500.00\\
Median rollout (rollout 0, reward=500.0000, length=500)
\end{quote}
 
\noindent\textbf{Critic-LLM reasoning:}
\begin{quote}\small\ttfamily
params[0]: 6, params[1]: 6, params[2]: 6, params[3]: 6, params[4]: -1, params[5]: 6, params[6]: -0.3, params[7]: 6, params[8]: -1.5, params[9]: -2
 
The collected statistics show a mean reward of 436.05 with a 75\% success rate (15/20 rollouts reached the full 500 step). \hl{The failures are sporadic;} \hl{the median rollout (full length 500)} \hl{is the same as with the current weights}, indicating that the policy is essentially correct but occasionally pushes the cart too hard at the start of some episodes.
 
Inspecting the median trajectory, the cart velocity term (params[1] and params[3]) dominates the action decision together with a large bias difference between the two actions (params[8] and params[9]). When the initial cart velocity is slightly high, the action chosen can generate a large impulse that drives the cart beyond the allowed bounds early in the episode.
 
To temper this over-reaction we reduce the magnitude of the theta-velocity weight in the first action (params[6]). A smaller negative value (-0.3 instead of -0.5) keeps the swing behavior largely the same while slightly decreasing the corrective force when the pole's angular velocity is positive.
 
We also tighten the bias gap between the two actions (params[8] from -1.8 to -1.5). The difference of 0.5 still favours action 1 when the pole tilts to the right, but the reduced bias makes the policy a bit less aggressive, further preventing the cart from overshooting early.
 
\hl{All other parameters are kept unchanged} because they already fall within the high-performance region identified in previous trials. \hl{The adjustments are small (at most 1.0)} \hl{and target only the aspects that appear linked} \hl{to the occasional failures}, aiming to raise the mean reward closer to the target of approximately 500 without harming the already good median performance.
\end{quote}
 
Despite this careful, well-justified reasoning, the revision destroys the policy ($436.05 \to 173.30$). This demonstrates that even minimal, statistically-informed edits can fail due to LLM hallucination in numerical parameter output, and why selection remains essential.
 
\subsection*{Loop Diagnosis (FrozenLake, run 2, episode 26)}
 
\textbf{Initial reward:} 0.45 \quad \textbf{Revised reward:} 0.70 \quad $\Delta = +0.25$
 
\noindent\textbf{Stats given to Critic-LLM:}
\begin{quote}\small\ttfamily
Reward: mean=0.45, min=0.00, max=1.00\\
Episode length: mean=31.0, min=8, max=100\\
Success rate: 9/20 rollouts reached reward=1.00\\
Failure rate: 11/20 rollouts finished below reward=1.00\\
Median rollout (rollout 0, reward=0.0000, length=16, outcome=terminated before the rollout cap)
\end{quote}
 
\noindent\textbf{Critic-LLM reasoning:}
\begin{quote}\small\ttfamily
params[0]: 0, params[1]: 3, params[2]: 3, params[3]: 3, params[4]: 0, params[5]: 0, params[6]: 0, params[7]: 0, params[8]: 3, params[9]: 1, params[10]: 0, params[11]: 2, params[12]: 1, params[13]: 2, params[14]: 1, params[15]: 3

I examined the failure pattern in the median rollout: \hl{the agent repeatedly cycled between states 9 and 10,} \hl{never making progress toward state 15.} This is caused by the current action at state 10 (action 1) which chooses from \{10$\to$10, 10$\to$11, 10$\to$14\}. The 10$\to$10 transition is a self-loop, and 10$\to$11 leads straight into a hole. Consequently, many episodes terminate early due to this loop rather than stochasticity.

To mitigate this, I changed the action at state 10 to action 0 (down-left-up), which redistributes transitions to \{10$\to$14, 10$\to$9, 10$\to$6\}. This breaks the 9$\leftrightarrow$10 loop and provides a higher probability of moving toward the goal (through state 14). All other state-action pairs remain unchanged because they already contribute positively or do not manifest systematic failures, as seen in earlier trials (e.g., trial 21 had a similar configuration with a mean reward of 0.75). This targeted change should lift the mean reward toward the 1.0 target without altering well-behaving parts of the policy.
\end{quote}

\section{Custom environment specifications}
\label{app:custom-envs}

To enable a direct comparison with prior work, we reuse three custom 
environments introduced by~\citep{zhou2025prompted} (Maze, Nim, and Pong) 
without modification. We restate their specifications here for 
self-containment; the original definitions are due to~\citep{zhou2025prompted}.

\paragraph{Maze.}
\begin{itemize}[leftmargin=*,itemsep=2pt,topsep=2pt]
\item \textbf{Description:} The agent traverses a $3 \times 3$ grid from a 
fixed start to a fixed goal. The grid contains impassable walls.
\item \textbf{Observation space:} Discrete(9) -- the agent's position.
\item \textbf{Action space:} Discrete(4) -- \{Up, Down, Right, Left\}.
\item \textbf{Reward:} $+1$ for reaching the goal, $-0.011$ per step.
\item \textbf{Termination:} Goal reached, or 100 steps elapsed.
\end{itemize}

\paragraph{Nim.}
\begin{itemize}[leftmargin=*,itemsep=2pt,topsep=2pt]
\item \textbf{Description:} A game starting with 10 sticks. The agent and 
a rule-based optimal opponent alternate turns, removing 1--3 sticks each. 
The player forced to take the last stick loses. The agent moves first.
\item \textbf{Observation space:} Box(11,) -- integer count of remaining sticks.
\item \textbf{Action space:} Discrete(3) -- number of sticks to remove (1, 2, or 3).
\item \textbf{Reward:} $+1$ for winning, $-1$ for losing.
\item \textbf{Termination:} All sticks removed.
\end{itemize}

\paragraph{Pong.}
\begin{itemize}[leftmargin=*,itemsep=2pt,topsep=2pt]
\item \textbf{Description:} The agent controls the left paddle. The right 
paddle is controlled by a rule-based optimal policy that never misses. 
The agent's goal is to return the ball as many times as possible.
\item \textbf{Observation space:} Box(5,) -- ball position $(x, y)$, ball 
velocity $(v_x, v_y)$, and agent paddle's $y$ position.
\item \textbf{Action space:} Discrete(3) -- \{Up, Stay, Down\}.
\item \textbf{Reward:} $+1$ each time the agent successfully hits the ball.
\item \textbf{Termination:} The agent misses the ball, or 3 successful hits.
\end{itemize}

\section{R2PO Prompt Templates}
\label{app:prompts}

We provide the prompts used by the Search-LLM and the Critic-LLM. The Search-LLM prompt closely follows the ProPS prompt design~\citep{zhou2025prompted}, which we adopt as the global-optimizer stage of our two-stage framework. Our novel contribution is the Critic-LLM prompt, which introduces a calibrated trajectory evidence package (median rollout, aggregate rollout statistics) and a revision rule controlled by $\tau_c$ that instructs the model to preserve already-strong policies. Separate variants are used for continuous (linear policy) and discrete (tabular policy) tasks. Placeholders in angle brackets (e.g., \runtimefield{<RANK>}, \runtimefield{<HISTORY>}) are filled at runtime.

The Critic-LLM prompt includes an environment description (\runtimefield{<ENV\_DESCRIPTION>}). We use the same environment descriptions as~\citep{zhou2025prompted} (Appendix~L), with minor corrections to typos and, in some cases, to the policy parameterization (e.g., the stated number of parameters).
The custom environments (Maze, Nim, and Pong) follow the specifications of \citep{zhou2025prompted}, which we restate for self-containment in Appendix~\ref{app:custom-envs}.

\subsection{Fixed-value placeholders}
\label{app:prompt-placeholders}

Two placeholders represent fixed optimization conventions held constant 
across all environments without per-task tuning:

\begin{itemize}[leftmargin=*,itemsep=2pt,topsep=2pt]
\item \runtimefield{<STEP\_SIZE>} $= 1.0$. Used in the continuous-policy 
Search-LLM prompt as a suggested exploration step, and in the 
continuous-policy Critic-LLM prompt as a soft upper bound on per-parameter 
adjustments unless evidence justifies a larger change. Not used by the 
discrete-policy prompts.

\item Continuous parameter range $[-6.0, 6.0]$ with one-decimal precision, 
hard-coded in both continuous prompts. This range comfortably contains 
the optimal parameter scales observed in preliminary runs across all 
continuous-state environments and is held fixed across all methods, 
including the ProPS and ProPS+ baselines, for matched-budget comparison.
\end{itemize}

\subsection{Search-LLM Prompt (Continuous)}

\begin{exbox}{Search-LLM: Continuous Prompt}
\raggedright
\ttfamily\footnotesize
You are good global optimizer, helping me find the global maximum of a mathematical\\
function $f(\mathrm{params})$. I will give you the function evaluation and the\\
current iteration number at each step. Your goal is to propose input values that\\
efficiently lead us to the global maximum within a limited number of iterations\\
(\runtimefield{<MAX\_ITERATIONS>}).\\[2pt]

\textbf{\# Regarding the parameters params:}\\
params is an array of \runtimefield{<RANK>} float numbers.\\
params values are in the range of [-6.0, 6.0] with 1 decimal place.\\[2pt]

\textbf{\# Here's how we'll interact:}\\
1. I will first provide MAX\_STEPS (\runtimefield{<MAX\_ITERATIONS>}) along with a few training examples.\\
2. You will provide your response in the following exact format:\\
\quad * Line 1: a new input `params[0]: , params[1]: , ..., params[\runtimefield{<RANK-1>}]:`\\
\qquad aiming to maximize the function's value $f(\mathrm{params})$.\\
\qquad Please propose params values in the range [-6.0, 6.0], with 1 decimal place.\\
\quad * Line 2: detailed explanation of why you chose that input.\\
3. I will then provide the function's value $f(\mathrm{params})$ at that point,\\
\quad and the current iteration.\\
4. We will repeat steps 2--3 until we reach the maximum number of iterations.\\[2pt]

\textbf{\# Remember:}\\
1. Do not propose previously seen params.\\
2. The global optimum should be around \runtimefield{<OPTIMUM>}. If you are below that,\\
\quad this is just a local optimum. You should explore instead of exploiting.\\
3. Search both positive and negative values. During exploration, use search step size\\
\quad of \runtimefield{<STEP\_SIZE>}.\\[2pt]

\textbf{\# Previous optimization history}\\
For reference, here are previously tried policies and their outcomes in a compact format.\\
Next, you will see examples of params and $f(\mathrm{params})$ pairs.\\
\runtimefield{<HISTORY>}\\[2pt]

Now you are at iteration \runtimefield{<STEP\_NUMBER>} out of\\
\runtimefield{<MAX\_ITERATIONS>}. Please provide the results in the indicated format.\\
Do not provide any additional texts.
\end{exbox}

\subsection{Search-LLM Prompt (Discrete)}

\begin{exbox}{Search-LLM: Discrete Prompt}
\raggedright
\ttfamily\footnotesize
You are good global optimizer, helping me find the global maximum of a mathematical\\
function $f(\mathrm{params})$. I will give you the function evaluation and the\\
current iteration number at each step. Your goal is to propose input values that\\
efficiently lead us to the global maximum within a limited number of iterations\\
(\runtimefield{<MAX\_ITERATIONS>}).\\[2pt]

\textbf{\# Regarding the parameters params:}\\
params is an array of \runtimefield{<RANK>} int numbers.\\
params values should be an integer chosen from \runtimefield{<ACTIONS>}.\\[2pt]

\textbf{\# Here's how we'll interact:}\\
1. I will first provide MAX\_STEPS (\runtimefield{<MAX\_ITERATIONS>}) along with a few training examples.\\
2. You will provide your response in the following exact format:\\
\quad * Line 1: a new input `params[0]: , params[1]: , ..., params[\runtimefield{<RANK-1>}]:`\\
\qquad aiming to maximize the function's value $f(\mathrm{params})$.\\
\qquad Please propose params values from \runtimefield{<ACTIONS>}.\\
\quad * Line 2: detailed explanation of why you chose that input.\\
3. I will then provide the function's value $f(\mathrm{params})$ at that point,\\
\quad and the current iteration.\\
4. We will repeat steps 2--3 until we reach the maximum number of iterations.\\[2pt]

\textbf{\# Remember:}\\
1. Do not propose previously seen params.\\
2. The global optimum should be around \runtimefield{<OPTIMUM>}. If you are below that,\\
\quad this is just a local optimum. You should explore instead of exploiting.\\
3. Search all the possible values of params.\\[2pt]

\textbf{\# Previous optimization history}\\
For reference, here are previously tried policies and their outcomes in a compact format.\\
Next, you will see examples of params and $f(\mathrm{params})$ pairs.\\
\runtimefield{<HISTORY>}\\[2pt]

Now you are at iteration \runtimefield{<STEP\_NUMBER>} out of\\
\runtimefield{<MAX\_ITERATIONS>}. Please provide the results in the indicated format.\\
Do not provide any additional texts.
\end{exbox}

\subsection{Critic-LLM Prompt (Continuous)}

\begin{exbox}{Critic-LLM: Continuous Prompt}
\raggedright
\ttfamily\footnotesize
You are an optimization agent, tasked to optimize a linear policy for a reinforcement\\
learning control task. An initial set of parameters was just proposed and tested.\\
Your job is to review, reflect on the test results and improve the policy parameters.\\[2pt]

\textbf{\# Task description}\\
\runtimefield{<ENV\_DESCRIPTION>}\\[2pt]

\textbf{\# Policy structure}\\
The policy is a linear function with \runtimefield{<RANK>} floating-point parameters.\\
params values should be in the range [-6.0, 6.0] with 1 decimal place.\\
The episodic reward measures how well the policy performs.\\[2pt]

\textbf{\# What was proposed}\\
The optimizer suggested these parameters:\\
\runtimefield{<PROPOSED\_PARAMS>}\\[2pt]

\textbf{\# How it performed}\\
Average reward over evaluation rollouts: \runtimefield{<ACHIEVED\_REWARD>}\\
Target reward for this task: around \runtimefield{<OPTIMUM>}.\\[2pt]

\textbf{\# What actually happened during the test}\\
Below is a summary of  \runtimefield{<K>} rollout statistics, along with one median rollout as a \\ representative example.
\runtimefield{<TRAJECTORY\_SUMMARY>}\\[2pt]

\textbf{\# Previous attempts}\\
For context, here are previous parameter settings and outcomes in a compact format.\\
\runtimefield{<HISTORY\_TEXT>}\\[2pt]

\textbf{\# Your task}\\
Follow these steps before writing any parameters:\\
1. Use the reward and episode-length statistics to separate systematic failures\\
\quad from occasional ones.\\
2. Identify which state dimensions and action responses are associated with the\\
\quad representative median rollout.\\
3. Preserve behaviors that are consistent with the stronger aggregate statistics.\\
4. Revise only the parameters most responsible for instability or underperformance.\\[2pt]

\textbf{\# Important points of consideration:}\\
- Use the reward and length statistics to judge whether the policy is consistently\\
\quad weak or only occasionally failing.\\
- If success rate is already high, avoid overfitting to the representative trajectory.\\
- Use the median rollout as evidence of typical behavior.\\
- Keep changes focused.\\
- When achieved\_reward >= \runtimefield{<REVISION\_THRESHOLD>}, prefer zero\\
\quad changes or only one very small, strongly justified edit.\\
- Prefer adjustments of \runtimefield{<STEP\_SIZE>} or less per parameter unless the\\
\quad evidence strongly supports a larger change.\\
Your response must have exactly two parts, in this order:\\[2pt]

\textbf{FIRST LINE (parameters only, nothing else before this):}\\
params[0]: <value>, params[1]: <value>, ..., params[\runtimefield{<RANK-1>}]: <value>\\[2pt]

\textbf{REMAINING LINES (your reasoning):}\\
Explain what you changed and why.
\end{exbox}

\subsection{Critic-LLM Prompt (Discrete)}

\begin{exbox}{Critic-LLM: Discrete Prompt}
\raggedright
\ttfamily\footnotesize
You are an optimization agent, tasked to optimize a tabular policy for a\\
reinforcement learning task. An initial set of parameters was just proposed and\\
tested. Your job is to review, reflect on the test results and improve the policy\\
parameters.\\[2pt]

\textbf{\# Task description}\\
\runtimefield{<ENV\_DESCRIPTION>}\\[2pt]

\textbf{\# What is a policy here}\\
The policy is an array of \runtimefield{<RANK>} integers. Each entry params[i]\\
says which action the agent takes in state i. The possible actions are\\
\runtimefield{<ACTIONS>}.\\[2pt]

\textbf{\# What was proposed}\\
The optimizer suggested the following policy:\\
\runtimefield{<PROPOSED\_PARAMS>}\\[2pt]

\textbf{\# How it performed}\\
The policy was evaluated in the environment and achieved an average reward of\\
\runtimefield{<ACHIEVED\_REWARD>}.\\
The target reward for this task is around \runtimefield{<OPTIMUM>}.\\[2pt]

\textbf{\# What actually happened during the test}\\
Below is a summary of  \runtimefield{<K>} rollout statistics, along with one median rollout as a \\ representative example.
\\
\runtimefield{<TRAJECTORY\_SUMMARY>}\\[2pt]

\textbf{\# Previous optimization history}\\
For reference, here are previously tried policies and their outcomes in a compact format.\\
\runtimefield{<HISTORY\_TEXT>}\\[2pt]

\textbf{\# Your task}\\
Follow these steps before writing any parameters:\\
1. Use the reward and episode-length statistics to identify whether failures are\\
\quad consistent versus occasional.\\
2. Use the reward and episode-length statistics to judge whether failures are\\
\quad systematic or mostly due to stochasticity.\\
3. Locate the state-action decisions in the policy that most likely explain the\\
\quad representative median rollout and the aggregate statistics.\\
4. Revise only the parameters most responsible for the failures or missed\\
\quad opportunities.\\[2pt]

\textbf{\# Important points of consideration:}\\
- Use the reward and length statistics to distinguish consistent problems from\\
\quad occasional failures.\\
- If success rate is already high, avoid over-correcting based on a single trajectory.\\
- Use the median rollout as a representative example of typical behavior.\\
- Keep changes focused. If part of the policy is working, preserve those parameters.\\

- When achieved\_reward >= \runtimefield{<REVISION\_THRESHOLD>}, prefer zero\\
\quad changes or only one very small, strongly justified edit.\\
- Remember that each parameter value must be chosen from \runtimefield{<ACTIONS>}.\\
Your response must have exactly two parts, in this order:\\[2pt]

\textbf{FIRST LINE (parameters only, nothing else before this):}\\
params[0]: <value>, params[1]: <value>, ..., params[\runtimefield{<RANK-1>}]: <value>\\[2pt]

\textbf{REMAINING LINES (your reasoning):}\\
Explain what you changed and why.
\end{exbox}

\

\begin{ack}
This work was supported by the Dubai RDI Grant Program under
the project ’CogWaters’ (Project ID: 2025/DRDI0163).
% Use unnumbered first level headings for the acknowledgments. All acknowledgments

% Do {\bf not} include this section in the anonymized submission, only in the final paper. You can use the \texttt{ack} environment provided in the style file to automatically hide this section in the anonymized submission.
\end{ack}

% \newpage
% \input{checklist.tex}

\end{document}